%% file: 20101001-PAMI.Manuscript.tex
\documentclass[journal]{IEEEtran}
\newcommand{\algfiguretwocol}[4]{
\setcounter{gap}{0}
\renewcommand{\baselinestretch}{1}
\begin{figure*}[#2]
\begin{center}
\rule{#1 \linewidth}{2.5pt}
\parbox{#1 \linewidth}{
\medskip
#3
\medskip
}
\rule{#1 \linewidth}{1.5pt}
\end{center}
\caption{#4}
\end{figure*}
\renewcommand{\baselinestretch}{1.11111111111111111111111111111111}
}

\usepackage{cite}

\usepackage{graphicx}
\usepackage{epsfig}
\usepackage{color}
\usepackage{epsf}
\usepackage{placeins}
\usepackage{multirow}

\input algo.tex

\usepackage[cmex10]{amsmath}

\usepackage{algorithmic}

\usepackage[tight,footnotesize]{subfigure}

\begin{document}
%
\title{Competitive Analysis of Minimum-Cut Maximum Flow Algorithms in Vision Problems}
%
%
%

\author{Barak Fishbain,
	    Dorit S. Hochbaum
	    and Stefan Mueller}
	    
\thanks{B. Fishbain (barak@berkeley.edu) and D.S. Hochbaum (hochbaum@ieor.berkeley.edu) are with the Dept. of Industrial Engineering and Operational Research at the University of California, Berkeley, Etcheverry Hall, University of California Berkeley, CA 94720}
\thanks{S. Mueller (ste.mu@arcor.de) is with the Combinatorial Optimization \& Graph Algorithms group at the Technische Universitaet Berlin, Germany}
\thanks{Manuscript received date; revised date.}

%



\maketitle

\begin{abstract}
Rapid advances in image acquisition and storage technology underline the need for algorithms that are capable of solving large scale image processing and computer-vision problems. The minimum cut problem plays an important role in processing many of these imaging problems such as, image and video segmentation, stereo vision, multi-view reconstruction and surface fitting. While several min-cut/max-flow algorithms can be found in the literature, their performance in practice has been studied primarily outside the scope of computer vision. We present here the results of a comprehensive computational study, in terms of execution times and memory utilization, of the four leading published algorithms, which optimally solve the {\em s-t} cut and maximum flow problems: (i) Goldberg's and Tarjan's {\em Push-Relabel}; (ii) Hochbaum's {\em pseudoflow}; (iii) Boykov's and Kolmogorov's {\em augmenting paths}; and (iv) Goldberg's {\em partial augment-relabel}. Our results demonstrate that while the {\em augmenting paths} algorithm is more suited for small problem instances or for problems with short paths from $s$ to $t$, the {\em pseudoflow} algorithm, is more suited for large general problem instances and utilizes less memory than the other algorithms on all problem instances investigated.
\end{abstract}


%
\IEEEpeerreviewmaketitle

\section{Introduction}
\label{sec:intro} 
\IEEEPARstart{T}{he} {\em minimum cut} problem (min-cut) and its dual, the {\em maximum flow} problem (max-flow), are classical combinatorial optimization problems with applications in numerous areas of science and engineering (for a collection of applications of min-cut and max-flow see \cite{Ahuja1993}).

Rapid advances in image acquisition and storage technology have increased the need for faster image processing and computer-vision algorithms that require lesser memory while being capable of handling large scale imaging problems. The min-cut problem takes a prominent role in many of these imaging problems, such as image and video segmentation, \cite{502093,HochbaumPAMI2009}, co-segmentation \cite{Hochbaum2009a}, stereo vision \cite{Scharstein2001}, multi-view
reconstruction, \cite{Sinha2008,Snavely2006}, and surface fitting
\cite{10.1109/MCG.2007.68}.

Several min-cut/max-flow algorithms can be found in the combinatorial
optimization literature. However, their performance in practice has been
studied primarily outside the scope of computer vision. In this study we compare, in terms of execution times and memory utilization, the four leading published algorithms, which solve optimally the min-cut and max-flow problems within the scope of vision problems. The study consists of a benchmark of an extensive data set which includes standard and non-standard vision problems \cite{CVRG,SCV3DR}.

The algorithms compared within the scope of this study are: (i)
the {\em Push-Relabel}, PRF, algorithm devised by Goldberg and
Tarjan \cite{Goldberg1988}; (ii) the Hochbaum's {\em
pseudoflow} algorithm, HPF \cite{Hochbaum2008}; (iii) Boykov's
and Kolmogorov's {\em augmenting paths} algorithm, BK,
\cite{Boykov2004}; and (iv) Goldberg's {\em partial
augment-relabel}, PAR, algorithm \cite{Goldberg2008}.

The study of these algorithms within the scope of computer-vision was reported in \cite{Boykov2004,Goldberg2008}. The first, \cite{Boykov2004}, compares the BK algorithm only to PRF, and for a limited set of instances. The latter report,\cite{Goldberg2008}, used the same limited set of instances, and
compared PRF and PAR to HPF. The comparison provided in \cite{Goldberg2008} to HPF is however not valid, as it did not use the updated publicly available software. Here we provide, for the first time, a comprehensive review of all these algorithms and a detailed comparison of several aspects of their performance, including a breakdown of the run-times and memory requirements. The breakdown of the run-times for the different stages of the algorithm: initialization, minimum cut computation and flow recovery, 
is important as the logic of the software is allocated differently by these algorithms to these stages. For example, while the
initialization process in the BK and HPF algorithms only reads the problem file and initiate the corresponding graphs, the implementation of the PRF incorporates an additional logic into this stage, e.g. sorting the arcs of each node. This extends the execution time of the initialization phase, and as a result of the entire algorithm. While our experiments show that this time is significant, it was disregarded in the previous reports in which the initialization execution time was not considered as part of the algorithm's running times. In addition, for many computer-vision applications only the min-cut solution is of importance. Thus, there is no need to recover the actual maximum flow in order to solve the problem. The breakdown of the execution times allows to evaluate the performance of the algorithms for these relevant computations by taking into account only the initialization and minimum-cut times.

Our results demonstrate that while the BK algorithm is more suited for small problem instances or for problems with short paths from $s$ to $t$, the HPF algorithm, is more suited for large general problem instances and utilizes less memory than the other algorithms on all problem instances investigated.

The paper is organized as follows: Section \ref{sec:algs}
describes the algorithms compared in this study. The experimental setup is presented in Section \ref{Section:expSetup}, followed by the comparison results, which are detailed in Section \ref{Section:results}. Section \ref{sec:Conclusion} concludes the paper.

\subsection{A graph representation of a vision problem}
A vision problem is typically presented on an undirected graph $G=(V,E)$, where $V$ is the set of pixels and $E$ are the pairs of adjacent pixels for which similarity information is available.    The $4$-neighbors setup is a commonly used adjacency rule with each pixel having $4$ neighbors -- two along the vertical axis and two along the horizontal axis. This set-up forms a planar grid graph.  The $8$-neighbors arrangement is also used, but then the planarity
of the graph is no longer preserved, and complexity of various algorithms increases, sometimes significantly. Planarity is also not satisfied for $3$-dimensional images or video. In the most general case of vision problems, no grid structure can be assumed and thus the respective graphs are not planar. Indeed, the algorithms presented here do not assume any specific property of the graph $G$ - they work for general graphs.

The edges in the graph representing the image carry {\em
similarity} weights.  There is a great deal of literature on
how to generate similarity weights, and we do not discuss this
issue here. We only use the fact that similarity is inversely
increasing with the difference in attributes between the
pixels.  In terms of the graph, each edge $\{i,j\}$ is assigned
a similarity weight $w_{ij}$ that increases as the two pixels
$i$ and $j$ are perceived to be more similar.  Low values of
$w_{ij}$ are interpreted as dissimilarity. In some
contexts one might want to generate {\em dissimilarity} weights
independently. In that case each edge has two weights, $w_{ij}$
for similarity, and $\hat{w}_{ij}$ for dissimilarity.

\subsection{Definitions and Notation}
\label{subsec:notations} Let $G_{st}$ be a graph $(V_{st},
A_{st})$, where $V_{st}=V\cup\{s,t\}$ and  $A_{st} = A\cup A_s
\cup A_t$ in which $A_s$ and $A_t$ are the source-adjacent and
sink-adjacent arcs respectively.  The number of nodes
$|V_{st}|$ is denoted by $n$, while the number of arcs
$|A_{st}|$ is denoted by $m$. A flow $f =\{f_{ij}\}_{(i,j) \in
A_{st}}$ is said to be {\em feasible} if it satisfies
\begin{enumerate}
\item [(i)] Flow balance constraints: for each $j \in V$,
    $\sum_{(i,j)\in A_{st}} f_{ij} = \sum_{(j,k)\in A_{st}}
    f_{jk}$ (i.e., inflow($j$) = outflow($j$)), and
\item [(ii)] Capacity constraints: the flow value is
    between the lower bound and upper bound capacity of the
    arc, i.e.,  $\ell_{ij} \leq f_{ij} \leq u_{ij}$.  We
    assume henceforth w.l.o.g that $\ell_{ij} = 0$.
\end{enumerate}

The {\em maximum flow} or {\em max-flow} problem on a directed
capacitated graph with two distinguished nodes---a source and a
sink---is to find a feasible flow $f^*$ that maximizes the
amount of flow that can be sent from the source to the sink
while satisfying flow balance constraints and capacity
constraints.

A {\em cut} is a partition of nodes $S\cup T=V$ with $s\in
S,t\in T$. Capacity of a cut is defined by $u(S,T)=\sum_{i\in
S, j\in T, (i,j)\in A} u_{ij}$. The {\em minimum $s$-$t$ cut}
problem, henceforth referred to as the {\em min-cut} problem,
defined on the above graph, is to find a bi-partition of
nodes---one containing the source and the other containing the
sink---such that the sum of capacities of arcs from the source
set to the sink set is minimized. In 1956, Ford and Fulkerson \cite{Fulkerson1956} established the {\em max-flow min-cut theorem}, which states that the value of a max-flow in any network is equal to the value of a min-cut. 

Given a capacity-feasible flow, hence a flow that satisfies
(ii), an arc $(i,j)$ is said to be a {\em residual arc} if
$(i,j)\in A_{st}$ and $f_{ij} < u_{ij}$ or $(j,i)\in A_{st}$
and $f_{ji}>0$.   For $(i,j)\in A_{st}$, the residual capacity
of arc $(i,j)$  with respect to the flow $f$ is
$c_{ij}^f=u_{ij}-f_{ij}$, and the residual capacity of the
reverse arc $(j,i)$ is $c^f_{ji}=f_{ij}$. Let $A^f$ denote the
set of residual arcs with respect to flow $f$ in $G_{st}$ which consists
of all arcs or reverse arcs with positive residual capacity.

A {\em preflow} is a relaxation of a flow that satisfies
capacity constraints, but inflow into a node is allowed to
exceed the outflow.  The {\em excess} of a node $v\in V$ is the
inflow into that node minus the outflow denoted by $e(v) =
\sum_{(u,v)\in A_{st}} f_{uv} - \sum_{(v,w)\in A_{st}} f_{vw}$.
Thus a preflow may have nonnegative excess.

A {\em pseudoflow} is a flow vector that satisfies capacity
constraints but may violate flow balance in either direction
(inflow into a node needs not to be equal outflow).  A negative excess
is called a {\em deficit}.

\section{Min-cut / Max-flow Algorithms}
\label{sec:algs}
\subsection{The push-relabel Algorithm}
\label{subsec:PR}

In this section, we provide a sketch of a
straightforward implementation of the algorithm. For a more
detailed description, see \cite{Ahuja1993,Goldberg1988}.

Goldberg's and Tarajen's push-relabel algorithm \cite{Goldberg1988}, PRF, works with {\em preflows}, where a node with strictly positive excess is said to be {\em active}. Each node $v$ is assigned a label $\ell(v)$ that satisfies (i)~$\ell(t) = 0$, and (ii)~$\ell(u) \leq \ell(v)+1$ if $(u,v) \in A^f$. A residual arc $(u,v)$ is said to be {\em admissible}
if $\ell(u) = \ell(v)+1$.

Initially, $s$'s label is assigned to be $n$, while all other nodes are assigned a label of $0$. Source-adjacent arcs are saturated creating a set of source-adjacent active nodes (all other nodes have zero
excess).  An iteration of the algorithm consists of selecting
an active node in $V$, and attempting to push its excess to its
neighbors along an admissible arc.  If no such arc exists, the
node's label is increased by $1$. The algorithm terminates with
a maximum preflow when there are no active nodes with
label less than $n$.  The set of nodes of label $n$ then forms
the source set of a minimum cut and the current preflow is
maximum in that it sends as much flow into the sink node as
possible. This ends Phase $1$ of the algorithm. In Phase $2$,
the algorithm transforms the maximum preflow into a maximum
flow by pushing the excess back to $s$. In practice, Phase $2$ is much faster than Phase $1$. A high-level description of the PRF algorithm is shown in Figure \ref{AlgFigure:pr}.

\algfiguretwocol{0.9}{htb}{ \Comment{Generic
push-relabel algorithm for maximum flow.}\nn \Procedure push-relabel$(V_{st}, A_{st}, c)$: \nn
  \Begin \nn
       Set the label of $s$ to $n$ and that of all other nodes to $0$;\nn
       Saturate all arcs in $A_s$;\nn
       \While there exists an active node $u \in V$ of label less
than $n$ \Do \nn
            \If there exists an admissible arc $(u,v)$ \Do \nn
                 Push a flow of $\min \{e(u), c^f_{uv}\}$ along arc
$(u,v)$; \EndLoop \nn
            \Else \Do \nn
                 Increase label of $u$ by $1$ unit; \EndLoop \EndLoop
\EndLoop \nn
  \End \EndLoop
}{\label{AlgFigure:pr}High-level description of Phase I of the generic push-relabel algorithm. The nodes with label equal to $n$ at termination form the source set of the minimum cut.}

The generic version of the PRF algorithm runs in
$O(n^2m)$ time. Using the dynamic trees data structure of
Sleator and Tarjan \cite{802464}, the complexity is improved to
$O(nm\log \frac{n^2}{m})$ \cite{Goldberg1988}. Two heuristics that are employed in practice significantly improve the run-time of the
algorithm: {\em Gap relabeling} and {\em Global relabeling} (see \cite{Goldberg1988,HPF-OR-2008} for details).

%

In the highest label and lowest label variants, an active node
with highest and lowest labels respectively are chosen for
processing at each iteration.  In the FIFO variant, the active
nodes are maintained as a queue in which nodes are added to the
queue from the rear and removed from the front for processing. In practice the FIFO - highest label variant is reported to work best\cite{Goldberg1988}. This variant of the algorithm is also referred to as HI\_PR. While, in this paper the highest label variant was used, it is referred to as PRF to indicate that this is the PRF algorithm. 

\subsection{The Hochbaum's Pseudo-flow Algorithm}
\label{subsec:HPF}

The Hochbaum's Pseudoflow algorithm, HPF, \cite{Hochbaum2008}
was motivated by an algorithm of Lerchs and Grossman \cite{LG1965}
for the maximum closure problem. The pseudoflow algorithm has
a strongly polynomial complexity of $O(nm \log \frac{n^2}{m})$ \cite{Hochbaum2009b}. Hochbaum's algorithm was shown to be fast in theory \cite{Hochbaum2008} and in practice \cite{HPF-OR-2008} for general benchmark problems.

Each node in $v \in V$ is associated with at most one {\em
current arc}, ${\rm currArc}(v) = (w,v)$, in $A^f$; the corresponding
{\em current node} of $v$ is denoted by ${\rm currNode}(v) =
w$. The algorithm also associates with each node with a {\em
root} that is defined constructively as follows: starting with
node $v$, generate the sequence of nodes $\{v, v_1, v_2, \dots,
v_r\}$ defined by the current arcs $(v_1, v), (v_2, v_1),
\dots, (v_r, v_{r-1})$ until $v_r$ has no current arc. Such
root node $v_r$ always exists \cite{HPF-OR-2008,Hochbaum2009b}. Let the unique root of node $v$ be denoted
by ${\rm root}(v)$.  Note that if node $v$ has no current
arc, then ${\rm root}(v) = v$.

The HPF algorithm is initiated with any arbitrary initial {\em pseudoflow} (i.e, flow vector that may violate flow balance in either direction) that saturates source adjacent and sink-adjacent arcs. Such initial pseudoflow can be generated, for example, by saturating all source-adjacent and sink-adjacent arcs, $A_s \cup A_t$, and setting all other arcs to have zero flow. This creates a set of source-adjacent nodes with excess, and a set of sink-adjacent nodes with deficit. All other arcs
have zero flow, and the set of initial current arcs is empty. Thus, each node is a singleton component of the forest for which it serves as a tree and the root of the tree.

The algorithm associates each node $v \in V$ with a distance
label $d(v)$.
A residual arc $(w,v)$ is said to be {\em admissible} if $d(w) = d(v)+1$.

A node is said to be {\em active} if it has strictly positive
excess. Given an admissible arc $(w,v)$ with nodes $w$ and $v$
in different components, an {\em admissible path} is the path from ${\rm root}(w)$ to ${\rm root}(v)$ along the
set of current arcs from ${\rm root}(w)$ to $w$, the arc
$(w,v)$, and the set of current arcs (in the reverse direction)
from $v$ to ${\rm root}(v)$.

An iteration of the HPF algorithm consists of choosing an
active component, with root node label $ < n$ and searching for an admissible arc from a {\em lowest labeled} node $w$ in
this component. Choosing a lowest labeled node for processing
ensures that an admissible arc is never between two nodes of
the same component.


By construction (see \cite{Hochbaum2008}), the root is the lowest labeled node in a component and node labels are non-decreasing with their distance from the root of the component. Thus, all the lowest labeled nodes within a component form a sub-tree rooted at the root of the component. Once an active component is identified, all the lowest labeled nodes within the component are examined for admissible arcs by performing a depth-first-search in the sub-tree starting at the root.

\begin{figure}[!h!t!b]
\epsfxsize = 0.9\linewidth
\centerline{\epsfbox{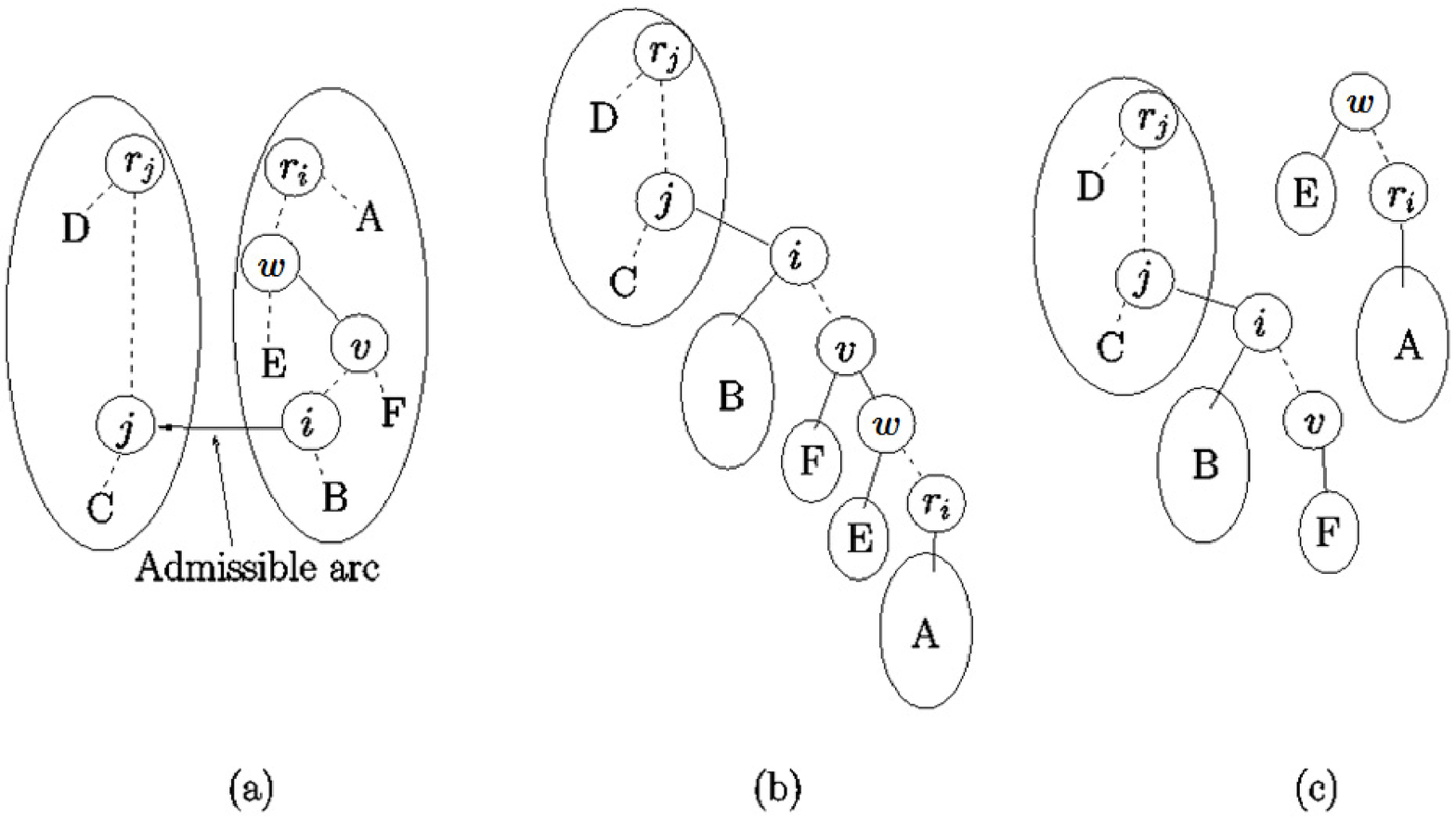}}
\caption{\label{fig:merger}(a) Components before merger (b)
Before pushing flow along admissible path from $r_i$ to $r_j$
(c) New components generated when arc $(w ,v)$ leaves the
current forest due to insufficient residual capacity.}
\end{figure}

If an admissible arc $(w,v)$ is found, a {\em merger} operation
is performed.  The merger operation consists of pushing the
entire excess of ${\rm root}(w)$ towards ${\rm root}(v)$ along
the admissible path and updating the excesses and the arcs in
the current forest. A schematic description of the merger
operation is shown in Figure \ref{fig:merger}. The pseudocode
is given in Figure \ref{AlgFigure:alt-monotone}.

If no admissible arc is found, $d(w)$ is increased by $1$ unit for all lowest label nodes $w$ in the component.
The algorithm terminates when there are no active
nodes with label $< n$. At termination all $n$ labeled nodes form the source set of the min-cut.

The active component to be processed in each iteration can be
selected arbitrarily. There are two variants
of the pseudoflow algorithm: (i) the lowest label pseudoflow
algorithm, where an active component with the lowest labeled
root is processed at each iteration; and (ii) the highest label
algorithm, where an active component with the highest labeled root
node is processed at each iteration.

The first stage of HPF terminates with the min-cut and a pseudoflow. The second stage converts this pseudoflow to a maximum feasible flow. This is done by {\em flow decomposition}. Hence representing the flow as the sum of flows along a set of $s$-$t$ paths, and flows along a set of directed cycles, such that no two paths or cycles are comprised of the same set of arcs (\cite{Ahuja1997}, pages 79-83). This stage can be done in $O(m \log n)$ by flow decomposition in a related network,  \cite{Hochbaum2008}.
Our experiments, like the experiments in \cite{HPF-OR-2008}, indicate that the time spent in flow recovery is small compared to the time to find the min-cut. 

\algfiguretwocol{0.9}{!h!t!b}{ {\Comment{Min-cut stage of HPF
algorithm.}}\nn \Procedure HPF ($V _{st},
A_{st}, c$): \nn
      \Begin \nn
      SimpleInit ($A_s, A_t, c$);\nn
      \While $\exists$ an active component $T$ with root $r$, where $d(r) < n$, \Do \nn
              $w \leftarrow r$;\nn
              \While $w \neq \emptyset$ \Do \nn
                   \If $\exists$ admissible arc $(w,v)$ \Do \nn
                           Merger (${\rm root}(w), \cdots, w,
v,\cdots,{\rm root}(v)$); \nn
                           $w \leftarrow \emptyset$; \EndLoop \nn
                   \Else \Do \nn
                           \If $\exists y\in T: ({\rm current}(y) =
w)  \wedge (d(y) = d(w))$ \Do \nn
                                $w \leftarrow y$;\EndLoop\nn
                           \Else \Do ~\{{\em relabel}\} \nn
                                $d(w) \leftarrow d(w) + 1$;\nn
                                $w \leftarrow {\rm
current}(w)$;\EndLoop \EndLoop \EndLoop \EndLoop \EndLoop \nn
      \End \EndLoop
}{\label{AlgFigure:alt-monotone} The min-cut stage of the HPF
algorithm. At termination all nodes in label-$n$ components are the source set of the min-cut.}

\subsection{Boykov's and Kolmogorov's Augmenting Paths Algorithm}
\label{subsec:BK} 

Boykov's and Kolmogorov's augmenting paths algorithm, BK, \cite{Boykov2004} attempts to improve on standard augmenting path techniques on graphs in vision. Given that $|C|$ is the capacity of a minimum cut, the theoretical complexity of this algorithm is $O(mn^2|C|)$. Similarly to Ford--Fulkerson's algorithm \cite{Fulkerson1956}, the BK algorithm's complexity is only pseudo-polynomial. In this it differs from the other algorithms studied here, all of which have strongly polynomial time complexity. Despite of that, it has been demonstrated in \cite{Boykov2004} that in practice on a set of vision problems, the algorithm works well. 

At heart of the augmenting paths approach is the use of search
trees for detecting augmenting paths from $s$ to $t$. Two such trees, one from the source, $T_S$, and the other from the sink, $T_T$ are
constructed, where $T_S \cap T_T = \emptyset$. The trees
are constructed so that in $T_S$ all edges from each parent node to
its children are non-saturated and in $T_T$, edges from
children to their parents are non-saturated.

Nodes that are not associated with a tree are called {\em free}. Nodes that are not free can be tagged as {\em active} or {\em passive}. Active nodes have edges to at least one free node, while passive nodes have no edges connecting them to a free node. Consequentially trees can grow only by connecting, through a non-saturated edge, a free node to an active node of the tree. An augmenting path is found when an active node in either of the trees detects a neighboring node that
belongs to the other tree.

At the initialization stage the search tree, $T_S$ contains only the source node, $s$ and the search tree $T_T$ contains only the sink node $t$. All other nodes are free.

Each iteration of the algorithm consists of the following three stages:

\noindent {\bf Growth} In this stage the search trees $T_S$ and $T_T$ expand. For all active nodes in a tree, $T_S$ or $T_T$,  adjacent free nodes, which are connected through non-saturated edge, are searched. These free nodes become members of the corresponding search tree. The growth stage terminates when the search for an active node from one tree, finds an adjacent (active) node that belongs to the other tree. Thus, an augmenting path from $S$ to $T$  was found.

\noindent {\bf Augmentation}  Upon finding the augmenting path, the maximum flow possible is being pushed from $s$ to $t$. This implies that at least one edge will be saturated. Thus, for at least one node in the trees $T_S$ and $T_T$ the edge connecting it to its parent is no longer valid. The augmentation phase may split the search trees $T_S$ and $T_T$ into forests. Nodes for which the edges connecting them to their parent become saturated are called {\em orphans}.

\noindent {\bf Adoption} In this stage the tree structure of $T_S$ and $T_T$ is restored. For each orphan, created in the previous stage, the algorithm tries to find a new valid parent. The new parent should belong to the same set, $T_S$ or $T_T$, as the orphan node and has a non-saturated edge to the orphan node. If no parent is found, then the  orphan node and all its children become free and the tree structure rooted in this orphan is discarded. This stage terminates when all orphan nodes are connected to a new parent or are free. 

The algorithm terminates when there are no more active nodes and the trees are separated by saturated edges. Thus, the maximum flow is achieved and the corresponding minimum-cut is $S=T_S$ and $T=T_T$.

It is interesting to note that there are two speed-ups for the BK-algorithm. The first one is an option to reuse search trees from one maxflow computation to the next as described in \cite{Kohli2005}. This option does not apply in our setting as the instances are not modified.
The other speed-up is due to capacity scaling \cite{Juan2007}. We would have liked to test this version but we are not aware of any publicly available implementation.

\subsection{The Partial Augment-Relabel}
\label{subsec:PAR} The Partial Augment-Relabel algorithm, PAR,
devised by Goldberg, \cite{Goldberg2008} searches for the shortest augmenting path and it maintains a flow (rather than a pseudoflow or preflow). A relabeling mechanism is utilized by the algorithm to find the augmenting paths.

The algorithm starts at $s$ and searches for admissible, non-saturated, arcs in a depth-first search manner. An arc $(x,y)$ is admissible if the label of its associated nodes is equal, $d(x) = d(y)$. At each iteration, the algorithm maintains a path from $s$ to $v\in V$ and tries to extend it. If $v$ has an admissible arc, $(v,w)$, the path is extended to $w$. If no such admissible arc is found, the algorithm shrinks the path,
making the predecessor of $v$ on the path the current node and
relabels $v$. At each iteration, the search terminates either if $w = t$, or if the length of the path reaches some predefined value, $k$, or if $v$, the current node has no outgoing admissible arcs. For $k = \Theta (\sqrt{m})$, PAR has a complexity of $O(n^2\sqrt{m})$
\cite{Goldberg2008}.

In order to achieve better performance in practice, the same gap and global heuristics mentioned in Section \ref{subsec:PR}, for PRF, can be applied here for the PAR algorithm.

\section{Experimental Setup}
\label{Section:expSetup}

The {\em PRF}, {\em HPF} and the {\em BK} algorithms are
compared here by running them on the same problem instances and on the same hardware setup. The run-times of the highest level variant of the PRF algorithm and of PAR are reported in \cite{Goldberg2008} for a subset of the problems used here. Since the source code for the PAR implementation is not made available, the PAR performance is evaluated here through the speedup factor of PAR with respect to the highest level variant of the PRF algorithm for each instance reported in the above paper.

As suggested by Chandran and Hochbaum \cite{HPF-OR-2008} we
use the highest label version for the HPF algorithm. The latest
version of the code (version 3.23) is available at
\cite{WebPS}. The highest level variant of the PRF algorithm is considered to have the best performance in practice \cite{Cherkassky1997}. We use the highest-level PRF implementation Version 3.5, \cite{WebHIPR}. Note that the latest implementation of the Push-Relabel method is actually denoted by HI\_PR, which indicates that the highest-label version is used. We refer to it as PRF, to indicate that it is the same algorithm which was reported in \cite{Cherkassky1997}. For the BK algorithm, a library implementation was used \cite{WebBK}. In order to utilize the library for solving problems in DIMACS format, a wrapping code, wrapper, was written. This wrapper reads the DIMACS files and calls the library's functions for constructing and solving the problem. The part that reads the DIMACS files, under the required changes, is similar to the code used in the HPF implementation. One should note that the compilation's setup and configuration of the library have great effect on the actual running times of the code. In our tests the shortest running times were achieved using the following compilation line {\tt g++ -w -O4 -o <output\_file\_name> -DNDEBUG -DBENCHMARK graph.cpp maxflow.cpp <wrapper\_implementation\_file>}.

Every problem instance was run $5$ times and we report the average time of the three runs. These are reported for the three different
stages of the algorithm (Initialization, Compute Min-Cut and
Flow recovery). As detailed in section \ref{sec:intro}, breaking down the run-times provides insight into the algorithms' performance and allows for better comparison. Since for many computer-vision applications only the min-cut solution is of importance (e.g. \cite{502093,HochbaumBioSig2007,Hochbaum2009a,Scharstein2001,Sinha2008,Snavely2006,10.1109/CVPR.1996.517099,Lempitsky2007,Ali2010,10.1109/CVPR.1999.784730,546886,Kwatra2005,Roy1998,Shi:2000zn,10.1109/CVPR.2000.854845}), the most relevant evaluation is of the initialization and min-cut times.

\subsection{Computing Environments}
Our experiments were conducted on a machine with x86\_64
Dual-Core AMD Opteron(tm) Processor at 2.4 GHz with 1024 KB
level 2 cache and 32 GB RAM. The operating system was GNU/Linux
kernel release 2.6.18-53.el. The code of all three algorithms, PRF, HPF and BK, was compile with gcc 4.1.2 for the x86\_64-redhat-linux platform with $-O4$ optimization flag.

One should note that the relatively large physical memory of
the machine allows one to avoid memory swaps between the memory
and the swap-file (on the disk) throughout the execution of the
algorithms. Swaps are important to avoid since when the machine's physical memory is small with respect to the problem's size, the memory swap operation might take place very often. These swapping times, the wait times for the swap to take place, can accumulate to a considerably long run-times. Thus, in these cases, the execution times are biased due to memory constraints, rather than measuring the algorithms' true computational efficiency. Therefore we chose large physical memory which allows for more accurate and unbiased evaluation of the execution times.

\subsection{Problem Classes}
\label{subsec:ProbClasses}
The test sets used consist of problem instances that arise as
min-cut problems in computer vision, graphics, and biomedical
image analysis. All instances were made available from the
Computer Vision Research Group at the University of Western
Ontario \cite{CVRG}. The problem sets used are classified
into four types of vision tasks: Stereo vision, Segmentation, Multi-view reconstruction; and Surface fitting. These are detailed in sections \ref{subsec:stereo} through \ref{subsec:surf}. The number of nodes $n$ and the number of arcs $m$ for each of the problems are given in Table
\ref{tab:probSizes}.

\subsubsection{Stereo Vision}
\label{subsec:stereo}
\begin{figure}[!h!t!b]
  \begin{center}
    \[\begin{array}{c c c}
	\framebox{\includegraphics[width=0.25\linewidth]{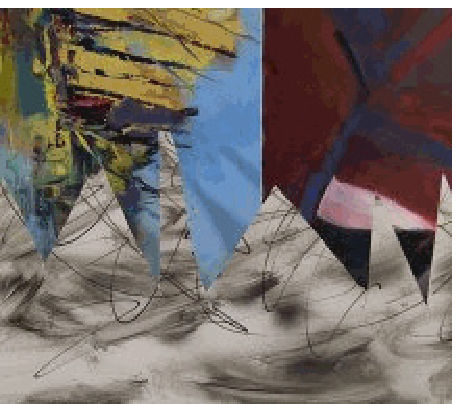}} &
      	\framebox{\includegraphics[width=0.25\linewidth]{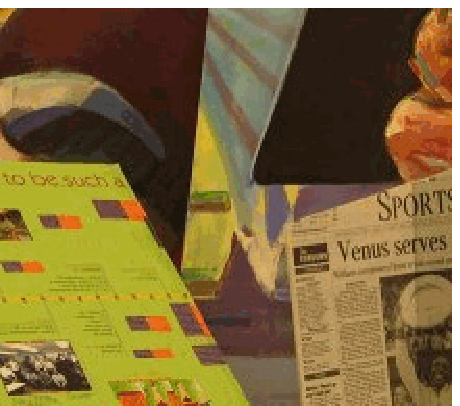}}&
      	\framebox{\includegraphics[width=0.295\linewidth]{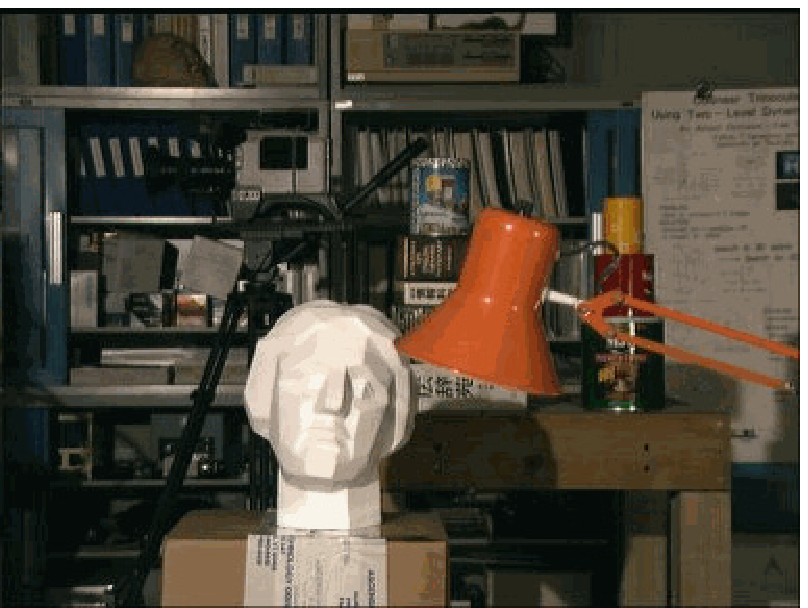}}\\
      Sawtooth & Venus & Tsukuba\\
    \end{array} \]
  \end{center}
  \caption{\label{Fig:StereoSeqs} Stereo test sequences (source \cite{CVRG})}
\end{figure}

Stereo problems, as one of the classical vision problems, have been extensively studied. The goal of stereo is to compute the correspondence between pixels of two or more images of the same scene. we use the {\em Venus}, {\em Sawtooth} \cite{Scharstein2001} and the {\em Tsukuba} \cite{10.1109/CVPR.1996.517099} data-sets. These sequences are made up of piecewise planar objects. Each of the stereo problems, used in this study, consists of an image sequence, where each image in the sequence is a slightly shifted version of its preceding one. A corresponding frame for each sequence is given in Figure \ref{Fig:StereoSeqs}.

Often the best correspondence between the pixels of the input images is determined by solving a min-cut problem for each pair of images in the set. Thus in order to solve the stereo problem, one has to solve a sequence of min-cut sub-problems all of approximately the same size. Previously reported run-times of these stereo problem \cite{Boykov2004,Goldberg2008} disclosed, for each problem, only the summation of the run-times of its min-cut sub-problems. Presenting the summation of the run-times of the sub-problems as the time for solving the entire problem assumes linear asymptotic behaviour of the run-times with respect to the input size. This assumption has not been justified. The run-times here, for the stereo problems, are reported as the {\em average} time it takes the algorithm to solve the min-cut sub-problem.

Each of the stereo min-cut sub-problems aims at matching  corresponding pixels in two images. The graphs consist of two $4$-neighborhood grids, one for each image. Each node, on every grid, has arcs connecting it to a set of nodes on the other grid. 
For each of the stereo problems there are two types of instances.  In one type, indicated by KZ2 suffix, each node in one image is connected to at most two nodes in the other image.  In the second type, indicated by BVZ suffix, each node in one image is connected to up to five nodes in the second image.

\subsubsection{Multi-view reconstruction}
\label{subsec:multiv}
\begin{figure}[!h!t!b]
  \begin{center}
    \[\begin{array}{c c}
      \framebox{\includegraphics[width=0.304\linewidth]{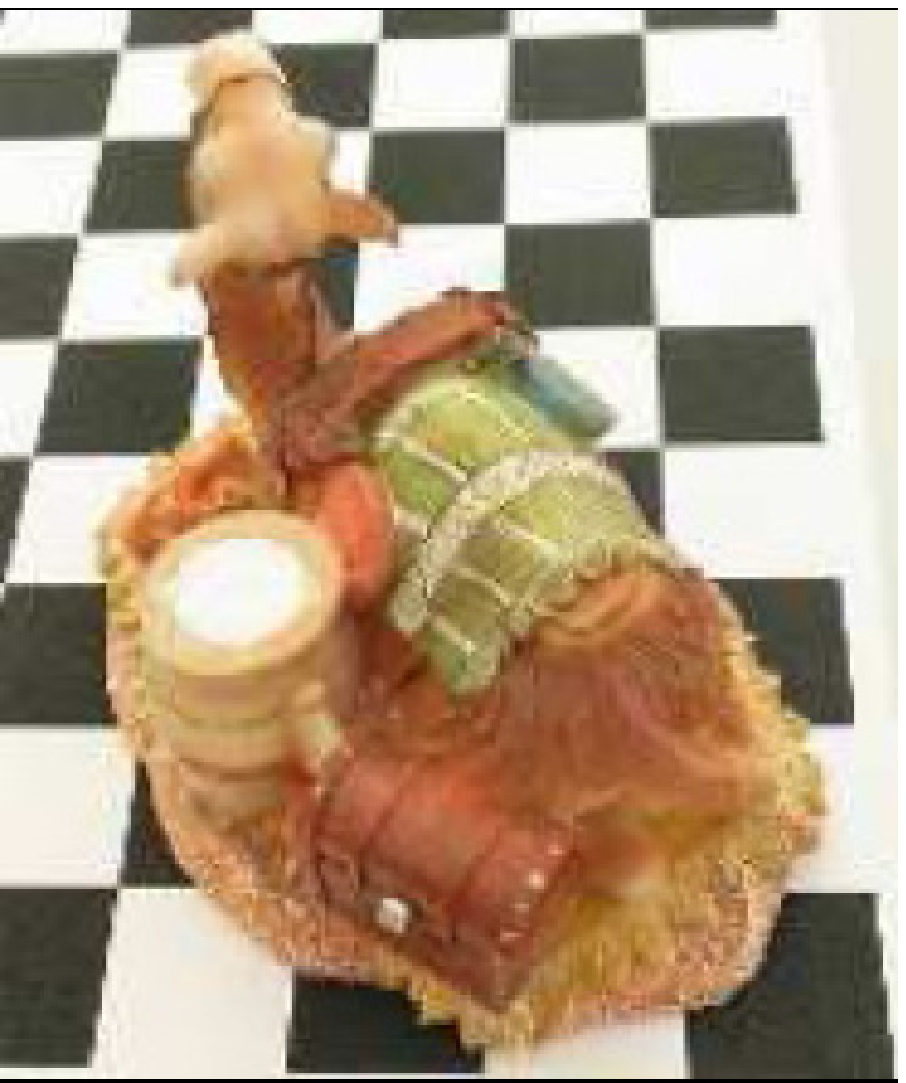}} &
      \framebox{\includegraphics[width=0.225\linewidth]{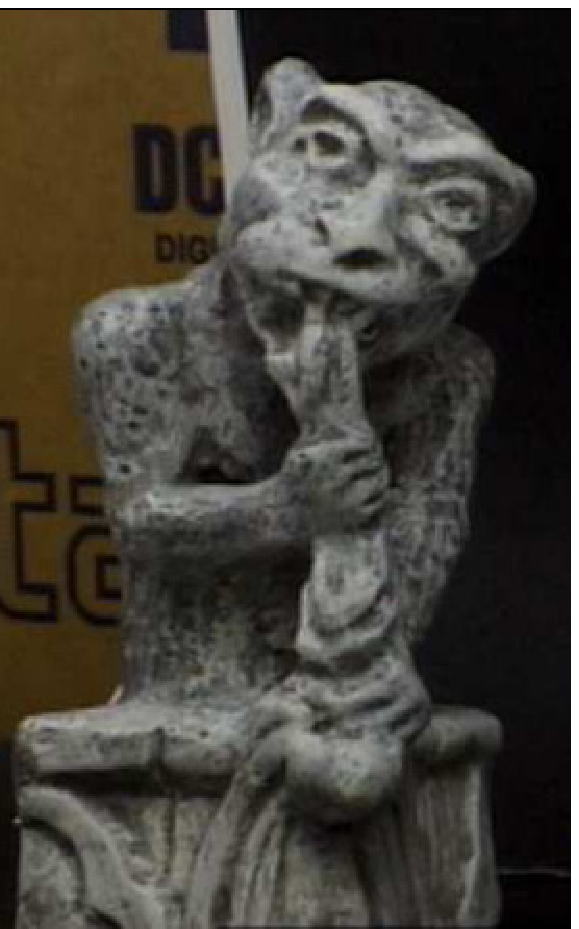}} \\
     Camel & Gargoyle\\
    \end{array} \]
  \end{center}
  \caption{\label{Fig:multiview} Multi-view test sequences (source \cite{CVRG})}
\end{figure}

A 3D reconstruction is a fundamental problem in computer vision with a
significant number of applications (for recent examples see      \cite{Ideses2007,Sinha2008,Snavely2006}). Specifically, graph theory based algorithms for this problem were reported in \cite{Lempitsky2006,Snow2000,Vogiatzis2005}.The input for the multi-view reconstruction problem is a set of 2D images of the same scene taken from different perspectives. The reconstruction problem is to construct a 3D image by mapping pixels from the 2D images to voxels complex in the 3D space. The most intuitive example for such a complex would be a rectangular grid, in which the space is divided into cubes. In the examples used here a finer grid, where 
each voxel is divided into 24 tetrahedral by six planes each passing through a pair of opposite cube edges, is used (See \cite{Lempitsky2006} for details). Two sequences are used in this class, {\em Camel} and {\em Gargoyle}. Each sequence was constructed in three different sizes (referred to as small, middle and large) \cite{YuriBoykov2006}. Representing frames are presented in Figure \ref{Fig:multiview}.

\subsubsection{Surface fitting}
\label{subsec:surf}
\begin{figure}[!h!t!b]
  \begin{center}
    \[\begin{array}{c c}
      \framebox{\includegraphics[width=0.304\linewidth]{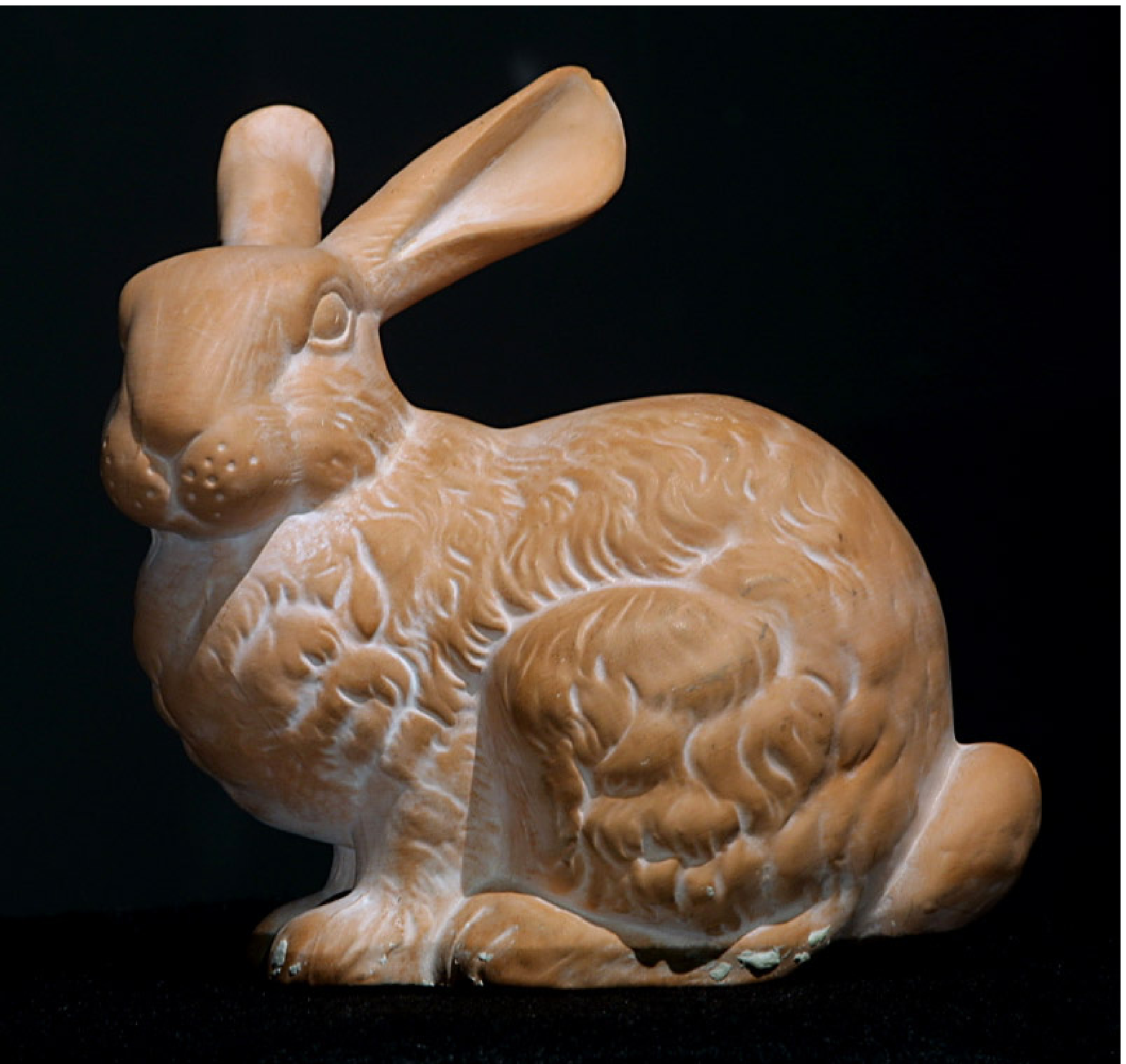}} &
      \framebox{\includegraphics[width=0.304\linewidth]{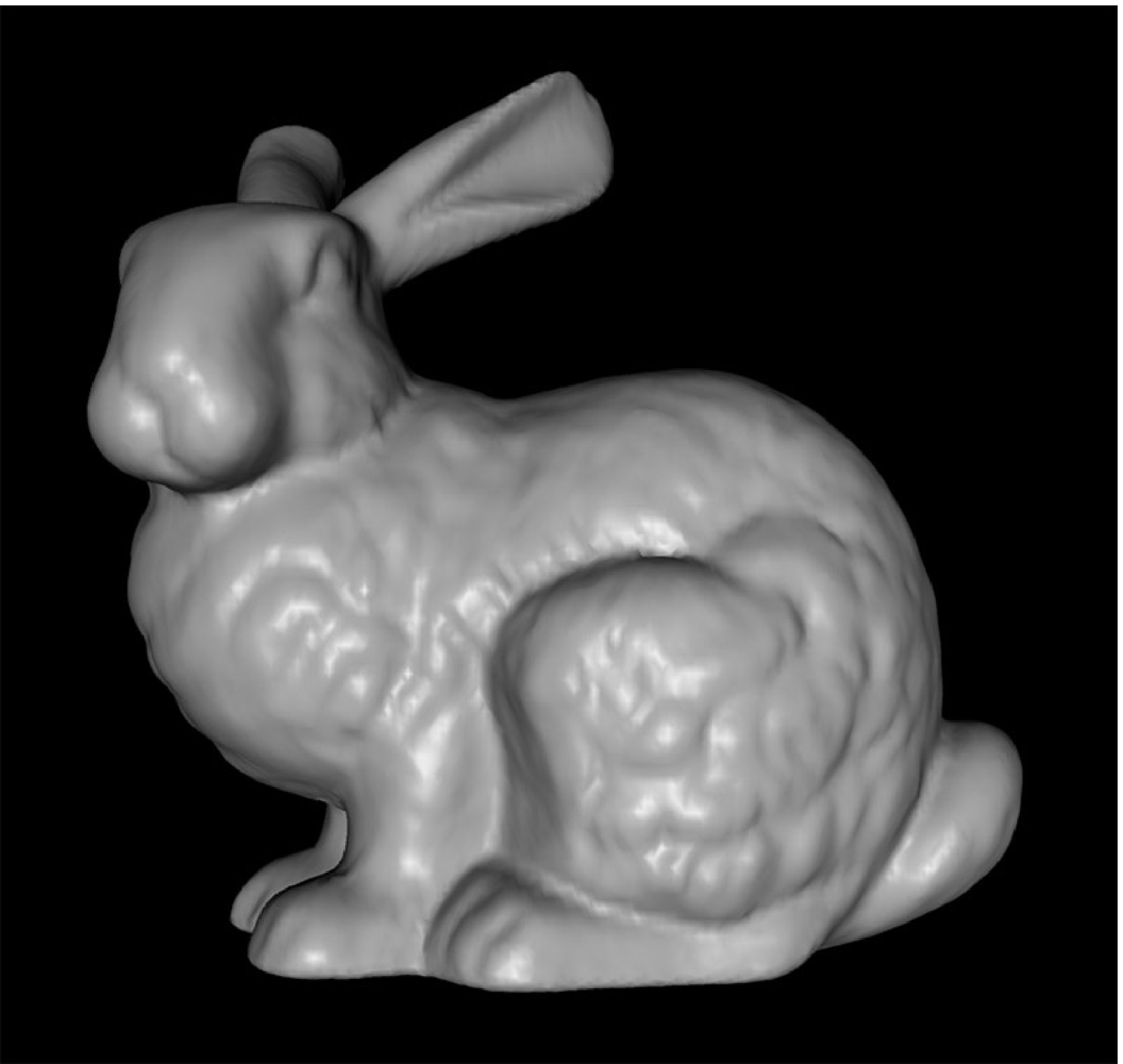}} \\
      Original Image & Reconstructed Surface \\
    \end{array} \]
  \end{center}
  \caption{\label{Fig:surffit} Bunny Problem Instance - Surface fitting (source \cite{SCV3DR})}
\end{figure}

3D reconstruction of an object's surface from sparse points containing noise, outliers, and gaps is also one of the most interesting problems in computer vision. Under this class we present a single test instance, {\em "Bunny"} (see Fig. \ref{Fig:surffit}), constructed in three different sizes. The sequence is part of the Stanford Computer Graphics Laboratory 3D Scanning Repository \cite{SCV3DR} and consists of $362,272$ scanned points. The goal then to reconstruct the observed object by optimizing a functional that maximizes the number of data points on the 3D grid while  imposing some shape priors either on the volume or the surface, such as spatial occupancy or surface area \cite{Lempitsky2007}. The "bunny" corresponding  graphs, on which the min-cut problem is solved, are characterized by particularly short paths from $s$ to $t$ \cite{Lempitsky2007}.

\subsubsection{Segmentation}
\label{subsec:seg}
Under this group $4$ test sets, referred to as {\em "Liver"}, {\em "adhead"}, {\em "Babyface"} and {\em "bone"} are used. Each set consists of similar instances which differ in the graph size,    neighborhood size, length of the path between $s$ and $t$, regional arc consistency (noise), and arc capacity magnitude \cite{Boykov2006}. For all instances used in this group, the suffices $n$ and $c$ represent the neighborhood type and maximum arc capacities respectively. For example, {\em bone.n6.c10} and {\em babyface.n26.c100}, correspond to a $6$ neighborhood and a maximum arc capacity of $10$ units and a $26$ neighborhood with maximum arc capacity of $100$ units respectively. The different {\em bone} instances differ in the number of nodes. The grid on the 3 axes x,y and z was made coarser by a factor of 2 on each, thus bone\_{\em xy}, means that the original problem (bone) was decimated along the x,y axes and it is $1/4$ of its original size; bone\_{\em xyz}, means that the original problem was decimated along the x,y and z axes and it is $1/8$ of its original size.

\begin{table*}[!h!t!b]
\begin{center}
\footnotesize
\begin{tabular}{||lrr|rrr|rrr||}
\hline \hline
\multicolumn{3}{||c|}{{\bf Instance}} & \multicolumn{3}{|c|}{{\bf Run-times [Secs]}} & \multicolumn{3}{c||}{{\bf Slowdown Factor}}\\ 
\hline
{Name} & {Nodes} & {Arcs} & {PRF} & {HPF} & {BK} &{PRF} & {HPF} & {BK}\\
\hline \hline
\multicolumn{9}{||l||}{{\bf Stereo}} \\ \hline
\input{table1ratio.tex}
\hline \hline
\end{tabular}
\caption{\label{tab:probSizes} Vision problems: Graph
sizes with combined Initialization and Minimum-cut run-times and their corresponding speedup factors. Each problem's fastest run-time is set in boldface. The speedup factor states how much an algorithm runs compared to the fastest algorithm}
\end{center}
\end{table*}

\section{Results}
\label{Section:results}

\subsection{Run-times}
\label{subsec:ResultsTimes}

In this study, the comparison of the PRF's, HPF's and BK's
run-times are indicated for the three stages of the algorithms:
(i) initialization, $t_{init}$; (ii) minimum-cut, $t_{minCut}$;
and (iii) maximum-flow, $t_{maxFlow}$. As these data
is unknown for the PAR algorithm, the comparison of these three algorithms with respect to PAR is addressed differently, by running PRF on our setup and deducing the PAR run-times by multiplying the measured PRF time by the speedup factor reported in \cite{Goldberg2008}. This is explained in Section \ref{subsec:PARComp}.

\begin{figure}[ht]

\begin{minipage}[t]{0.4\linewidth}
\centering
\subfiguretopcaptrue
\subfigure[Stereo]{
\frame{\includegraphics[width=\linewidth]{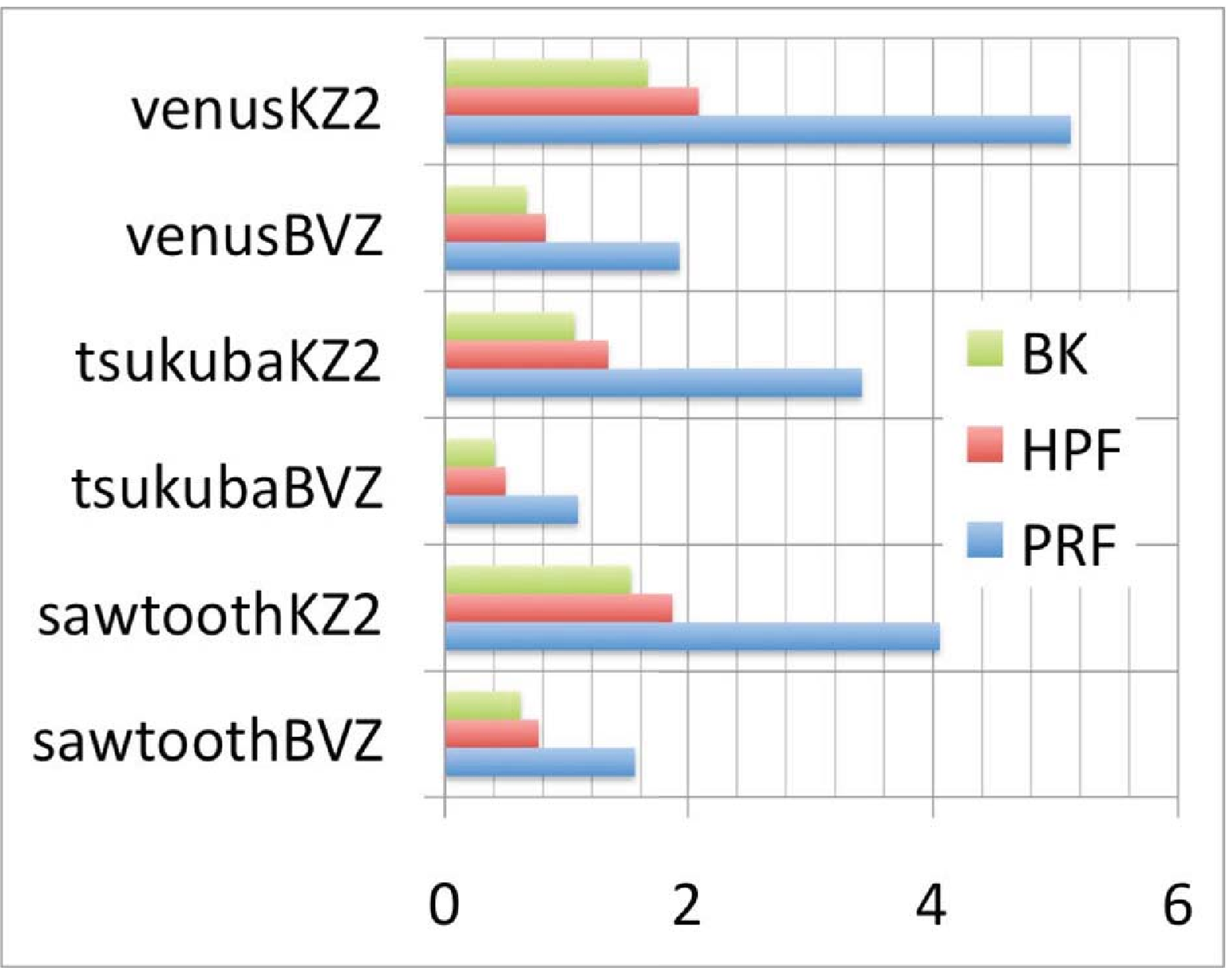}}
\label{fig:minStereo}
}
\subfigure[Multi-View]{
\frame{\includegraphics[width=\linewidth]{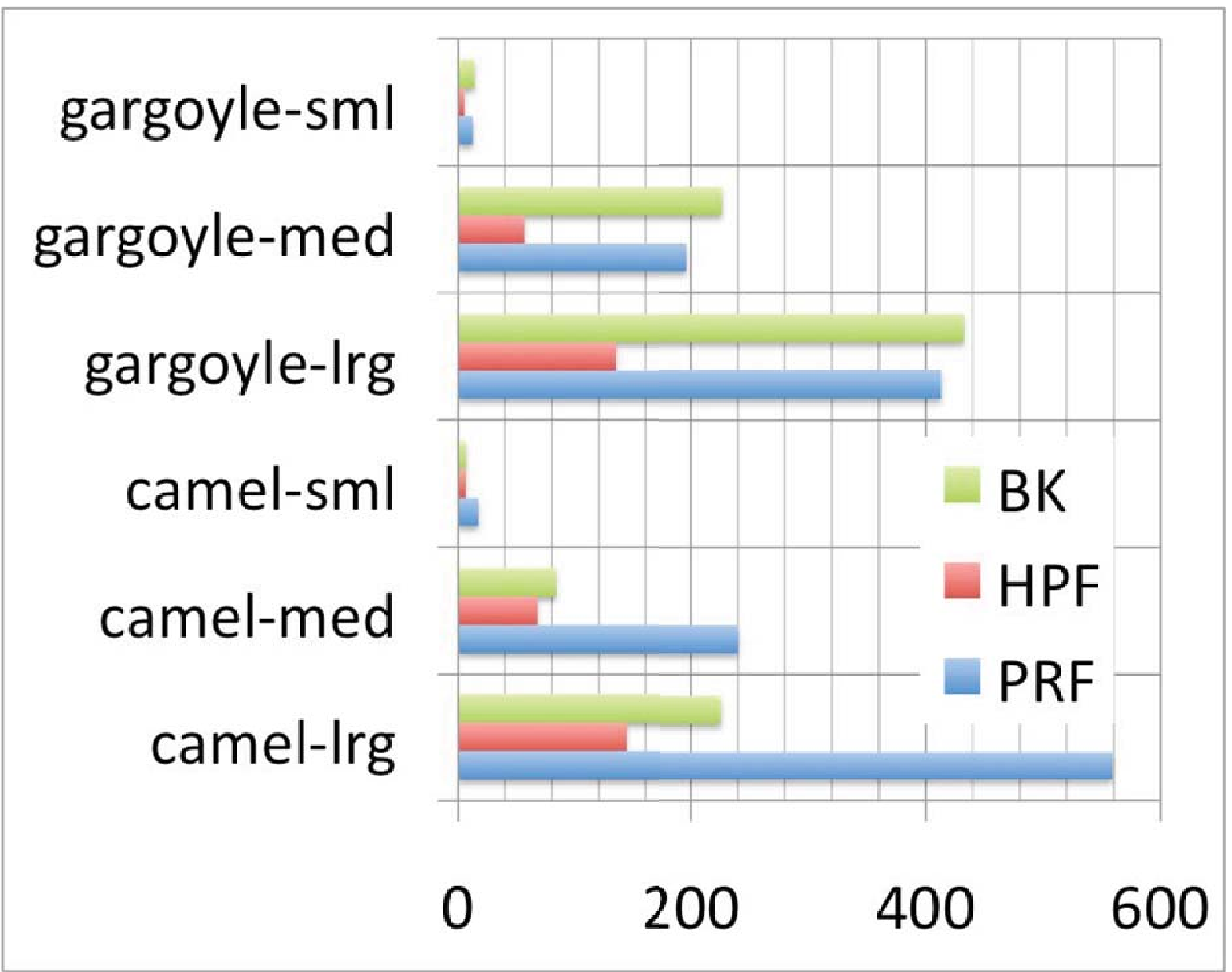}}
\label{fig:minMultiV}
}\\
\subfigure[Surface Fitting]{
\frame{\includegraphics[width=\linewidth]{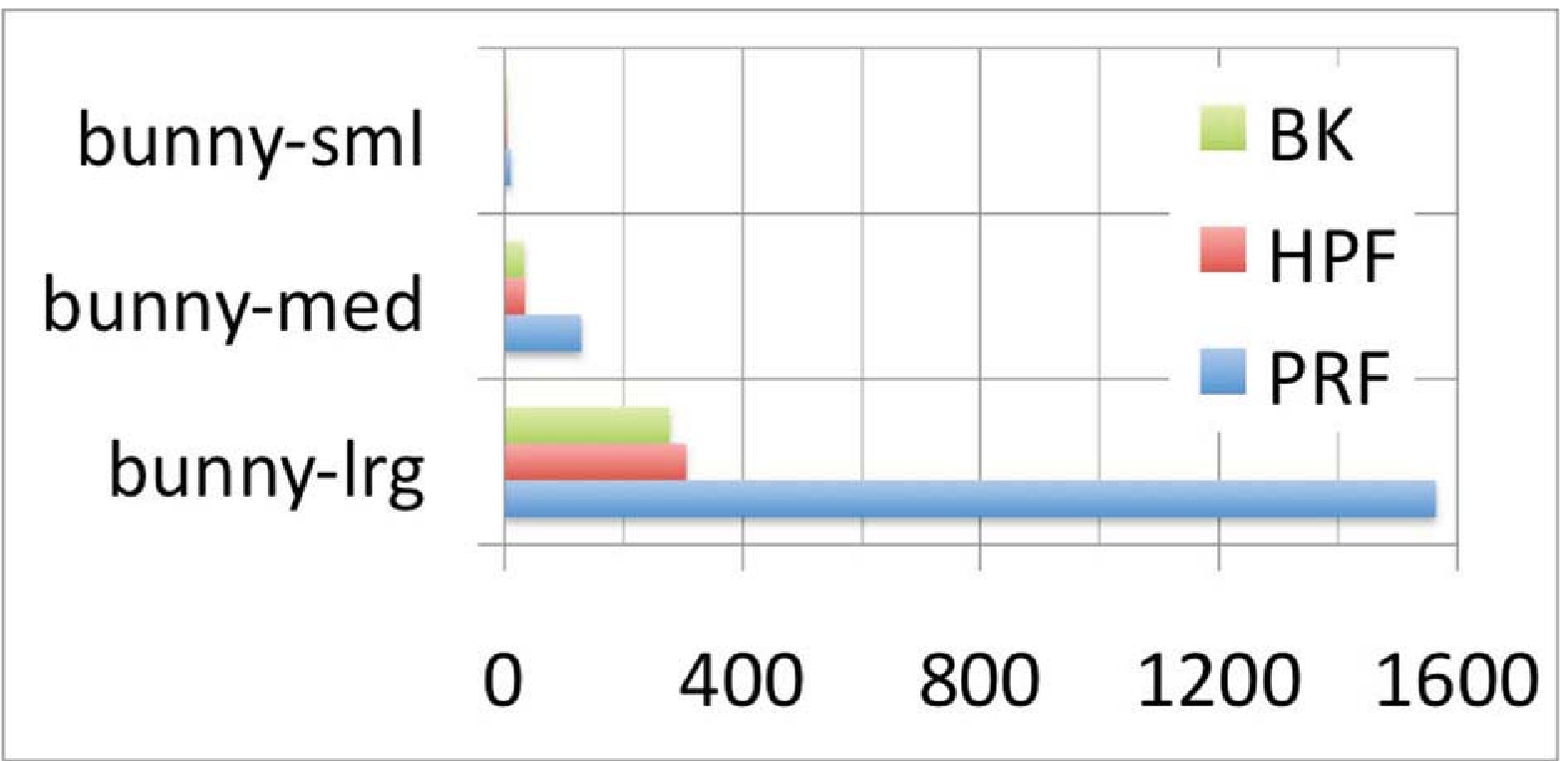}}
\label{fig:minSurf}
}
\end{minipage}
\hspace{0.01in}
\begin{minipage}[t]{0.5\linewidth}
\centering
\subfiguretopcaptrue
\subfigure[Segmentation]{
\frame{\includegraphics[width=\linewidth]{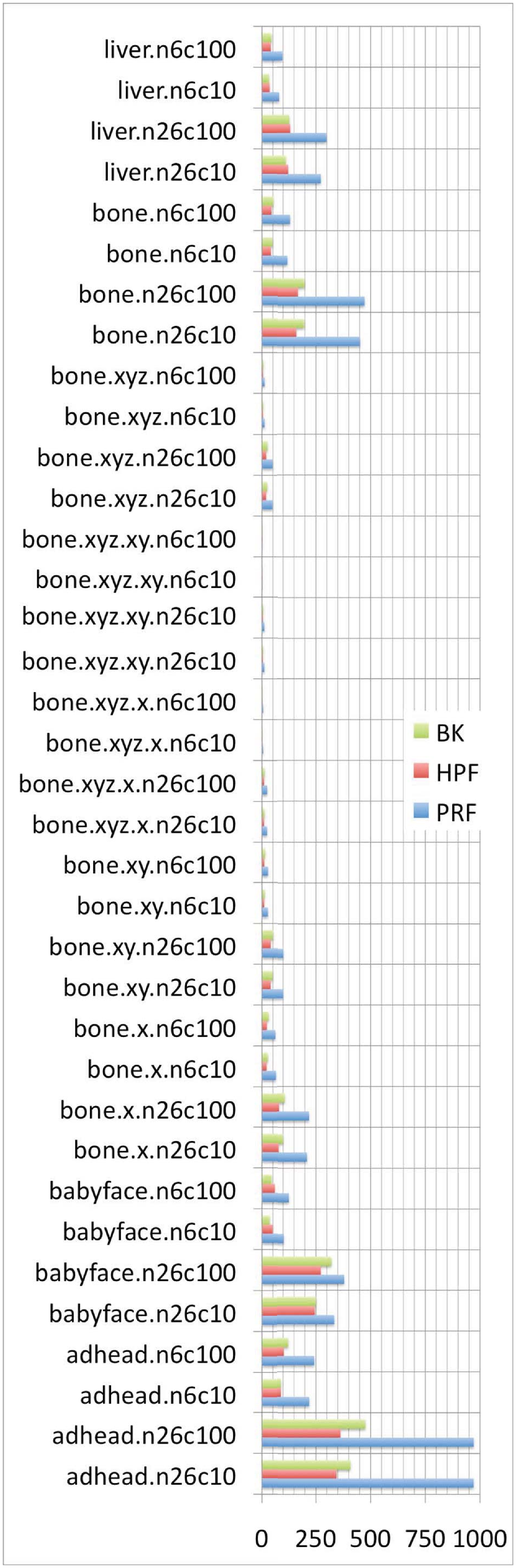}}
\label{fig:minSeg}
}
\end{minipage}
\caption{\label{fig:minTimes}{\bf Initialization and Minimum-cut} run-times in seconds: \subref{fig:minStereo} Stereo Problems; \subref{fig:minMultiV} Multi-view Problems; \subref{fig:minSurf} Surface Fitting; \subref{fig:minSeg} Segmentation}
\end{figure}

As already indicated, the most relevant times in this study are the times it takes each of the algorithms to complete the
computation of the min-cut, thus $t_{init} + t_{minCut}$. These are graphically presented in Figure \ref{fig:minTimes} and detailed in Table \ref{tab:probSizes}. The {\em Slowdown Factor}, reported in table \ref{tab:probSizes} for each algorithm, for every problem instance, is the ratio of the time it takes the algorithm to complete the computation of the minimum-cut divided by the minimum time it took any of the algorithms to complete this computation. 

Figure \ref{fig:minTimes}\subref{fig:minStereo} presents the run-times for the stereo vision problem sets. The input's size, for these problems is small, with respect the the other problem sets. For these small problem instances, the BK algorithm is doing better than PRF (with average Slowdown factor of $2.86$, which corresponds to average difference in the running time of $2.0$ Seconds) and slightly better than HPF (slowdown factor of $1.24$, which corresponds to a running time difference of $0.24$ Seconds). For the Multi-view instances HPF presents better results than both algorithms with average slowdown factors of $1.46$ with respect to BK and $3.19$ with respect to PRF. These correspond to differences in the running times of $95$ and $170$ seconds respectively. This is illustrated in Figure
\ref{fig:minTimes}\subref{fig:minMultiV}. Figure \subref{fig:minSurf} shows that the BK algorithm is more suitable for solving the surface fitting instances. This is attributed to the fact that these problems are characterized by particularly short s-t paths. In these instances, the slowdown factors of HPR and PRF are $1.05$ (correspond to an average difference of $9$ seconds) and $4.06$ (difference of $454$ seconds). The running times for the Segmentation problems class are depicted in Figure \ref{fig:minTimes}\subref{fig:minSeg}. There are $36$ segmentation problems. In a subset of $5$ segmentation problems BK achieved shorter running times. In this subset the BK's average slowdown factors are $1.19$ ($9.24$ seconds difference) and $2.62$ ($106$ seconds difference in the running time) with respect to HPF and PRF respectively. On the rest of the $31$ segmentation problems, HPF shows shorter running times with slowdown factors of $1.18$ ($14.22$ seconds difference) with respect to BK and $2.62$ ($101.39$ seconds difference) with respect to PRF.

A total of $51$ problem instances were tested within the scope of this study. The HPF algorithm was shown to be better in $37$ problem instances. The BK algorithm achieved better results over the other $14$ problems. The average run-times and slowdown factors of these two subsets are given in Table \ref{tab:aveResults}.

\begin{table}[htdp]
\caption{\label{tab:aveResults} Average running times and slowdown factors}
\begin{center}
\begin{tabular}{|c||c|c|c|}
\hline
& {\bf PRF} & {\bf HPF} & {\bf BK} \\
\hline
\multicolumn{4}{|l|}{{\bf HPF is better ($37$ problem instances)}}\\
\hline
Ave. run-time & 184.87 & 72.25 & 99.69\\
Ave. slowdown & 2.6 & 1 & 1.39 \\
\hline
\multicolumn{4}{|l|}{{\bf BK is better ($14$ problem instances)}}\\
\hline
Ave. run-time & 185.96 & 53.53 & 48.00\\
Ave. slowdown & 3.03 & 1.18 & 1 \\
\hline
\end{tabular}
\end{center}
\end{table}%

In order to allow for the comparison of the times it takes each
of the algorithms to complete the computation of only the
min-cut phase, the initialization run-times are presented in Figure \ref{fig:initTimes} and detailed in Appendix \ref{appen:runtimes}, Tables
\ref{tab:initstereo} -- \ref{tab:initseg}. Ideally one should be able to evaluate the minimum-cut processing times by subtracting the
initialization times in Tables \ref{tab:initstereo} --
\ref{tab:initseg} from the corresponding times in Table \ref{tab:probSizes}. However, as described
in Section \ref{sec:intro}, while the BK and HPF algorithms
only read the problem's data and allocate memory,
the PRF algorithm has some additional logic in its
initialization phase. Consequentially, one can not evaluate
PRF's min-cut processing times by this substraction. To
accomplish that, one has to account for the time it takes the
PRF algorithm to compute the additional logic implemented with
the initialization stage. Figure  \ref{fig:initTimes} shows that for all problem instances, the PRF's initialization times ($t_{init}$) are $2 - 3$ times  longer than BK's and HPF's times. While these times were excluded from the total execution times reported in \cite{Boykov2004} and \cite{Goldberg2008}, Figure \ref{fig:initTimes} strongly
suggests that these initialization times are significant with respect to
to the min-cut computation times ($t_{minCut}$) and should not be disregarded.

\begin{figure}[ht]

\begin{minipage}[t]{0.4\linewidth}
\centering
\subfiguretopcaptrue
\subfigure[Stereo]{
\frame{\includegraphics[width=\linewidth]{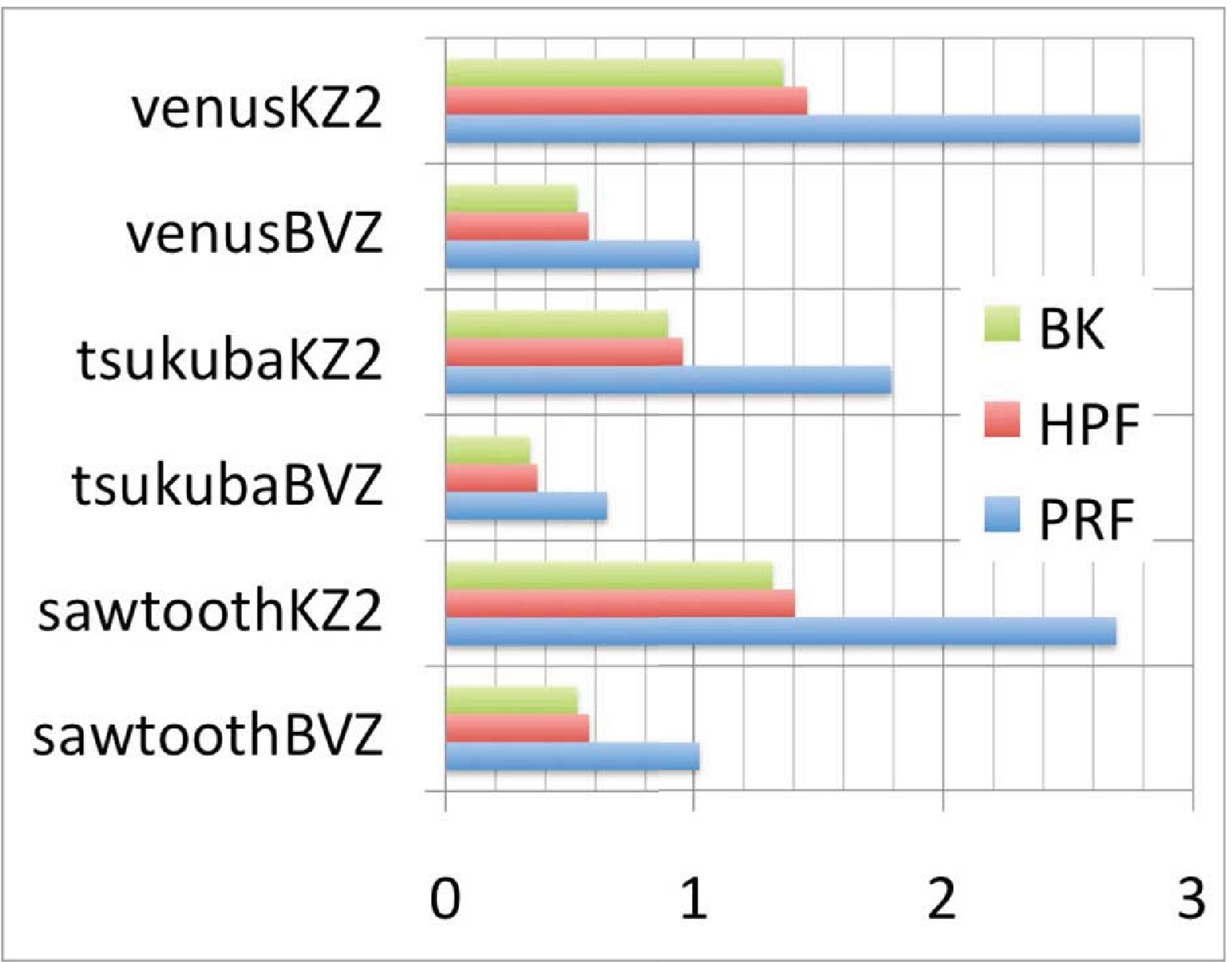}}
\label{fig:initStereo}
}
\subfigure[Multi-View]{
\frame{\includegraphics[width=\linewidth]{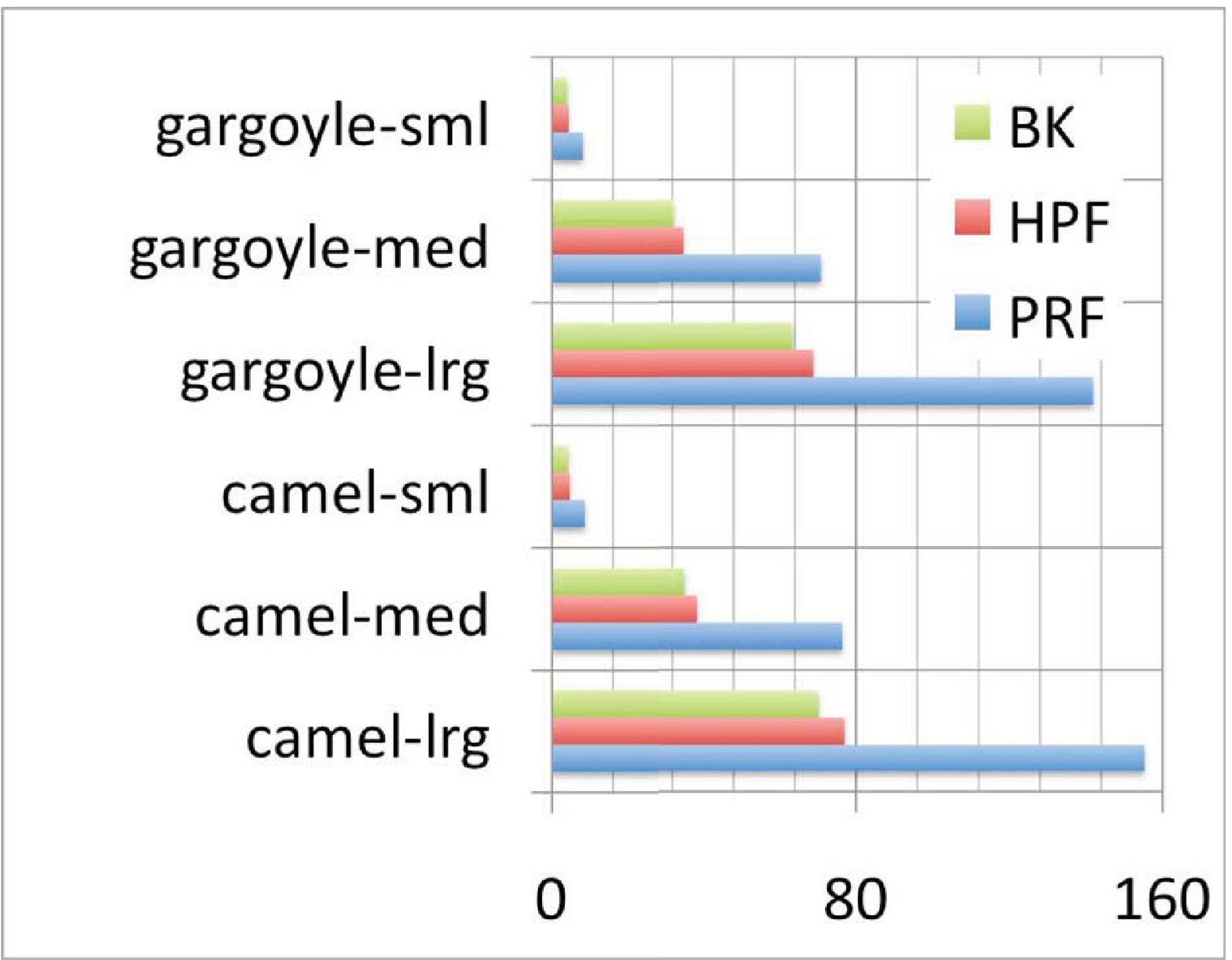}}
\label{fig:initMultiV}
}\\
\subfigure[Surface Fitting]{
\frame{\includegraphics[width=\linewidth]{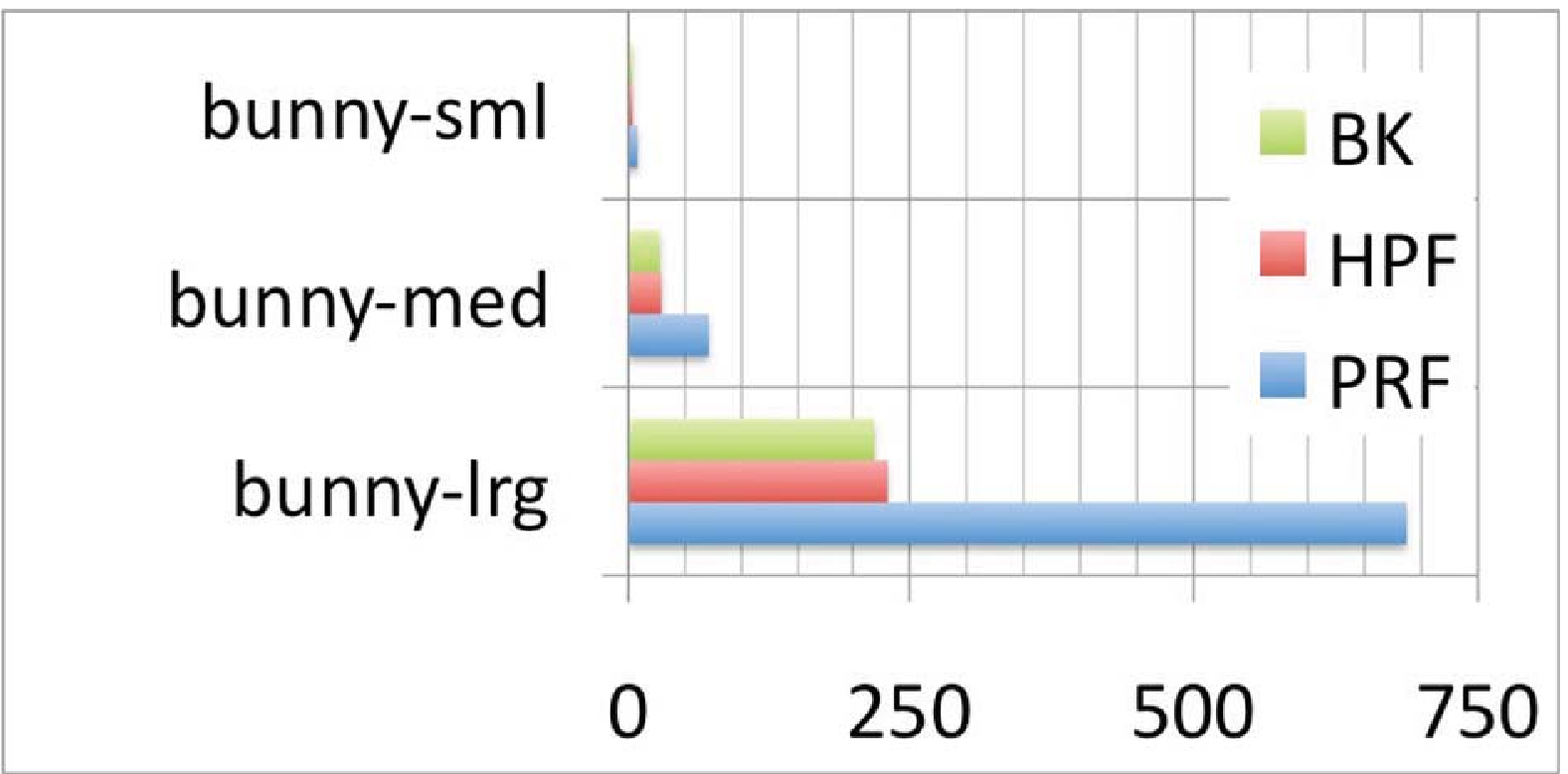}}
\label{fig:initSurf}
}
\end{minipage}
\hspace{0.01in}
\begin{minipage}[t]{0.5\linewidth}
\centering
\subfiguretopcaptrue
\subfigure[Segmentation]{
\frame{\includegraphics[width=\linewidth]{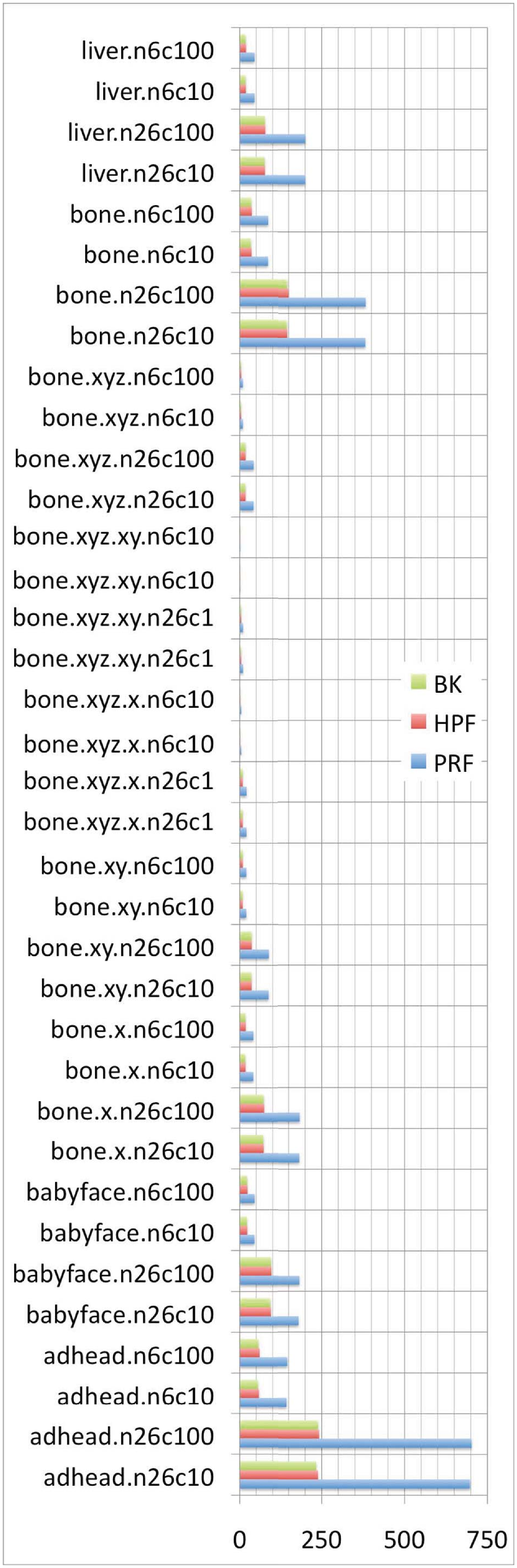}}
\label{fig:initSeg}
}
\end{minipage}
\caption{\label{fig:initTimes}{\bf Initialization} run-times in seconds: \subref{fig:initStereo} Stereo Problems; \subref{fig:initMultiV} Multi-view Problems; \subref{fig:initSurf} Surface Fitting; \subref{fig:initSeg} Segmentation}
\end{figure}

\begin{figure}[ht]

\begin{minipage}[t]{0.4\linewidth}
\centering
\subfiguretopcaptrue
\subfigure[Stereo]{
\frame{\includegraphics[width=\linewidth]{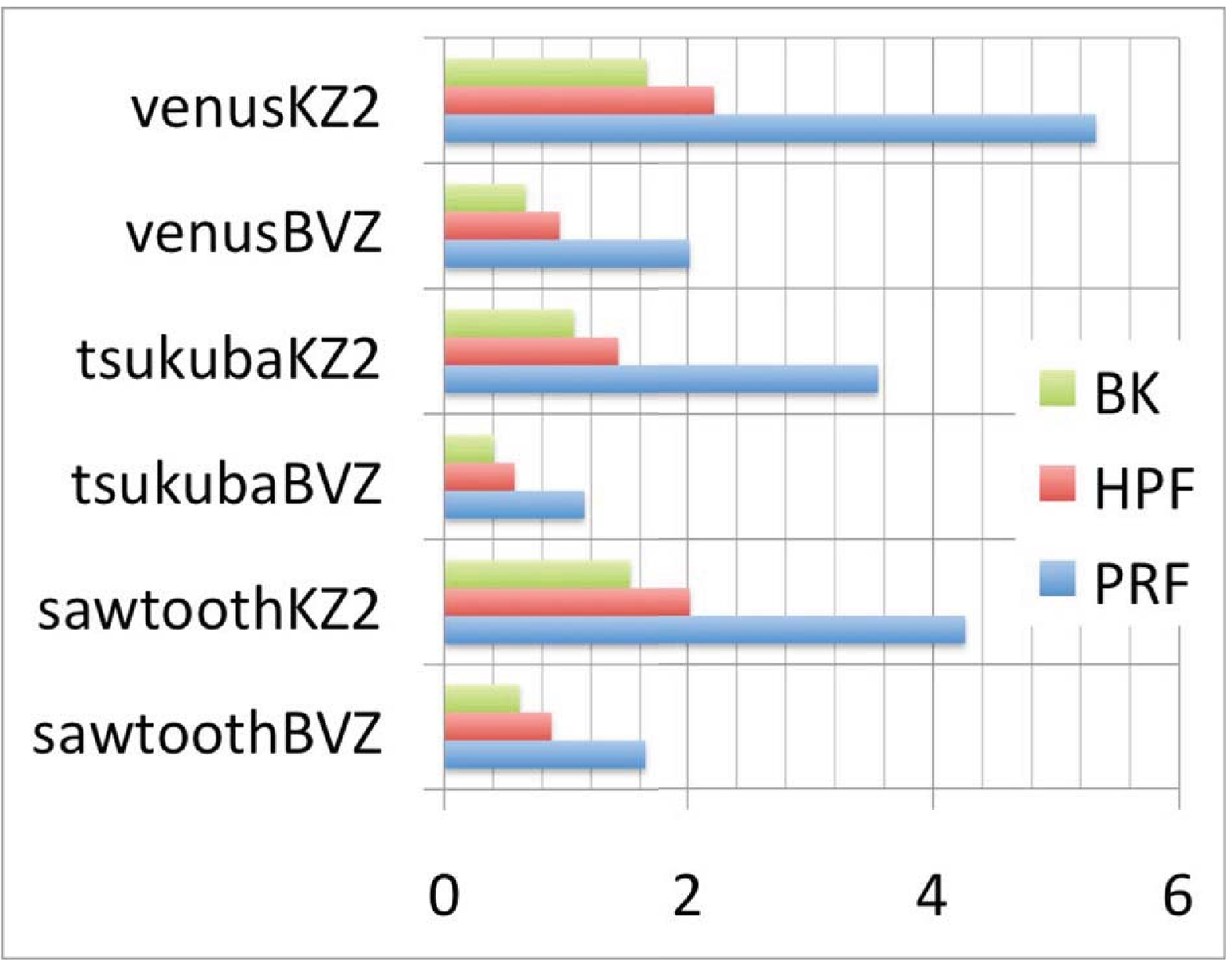}}
\label{fig:maxStereo}
}
\subfigure[Multi-View]{
\frame{\includegraphics[width=\linewidth]{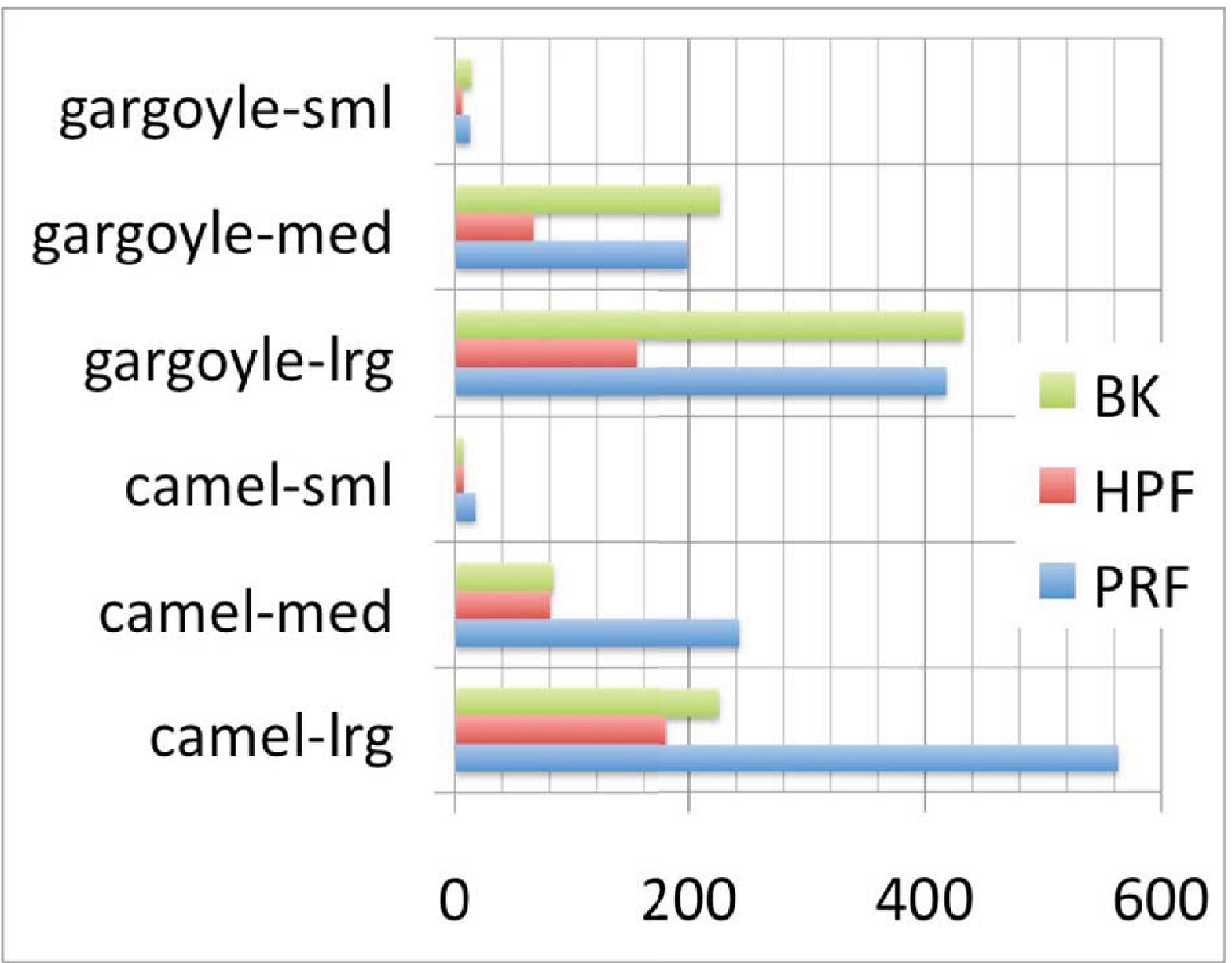}}
\label{fig:maxMultiV}
}\\
\subfigure[Surface Fitting]{
\frame{\includegraphics[width=\linewidth]{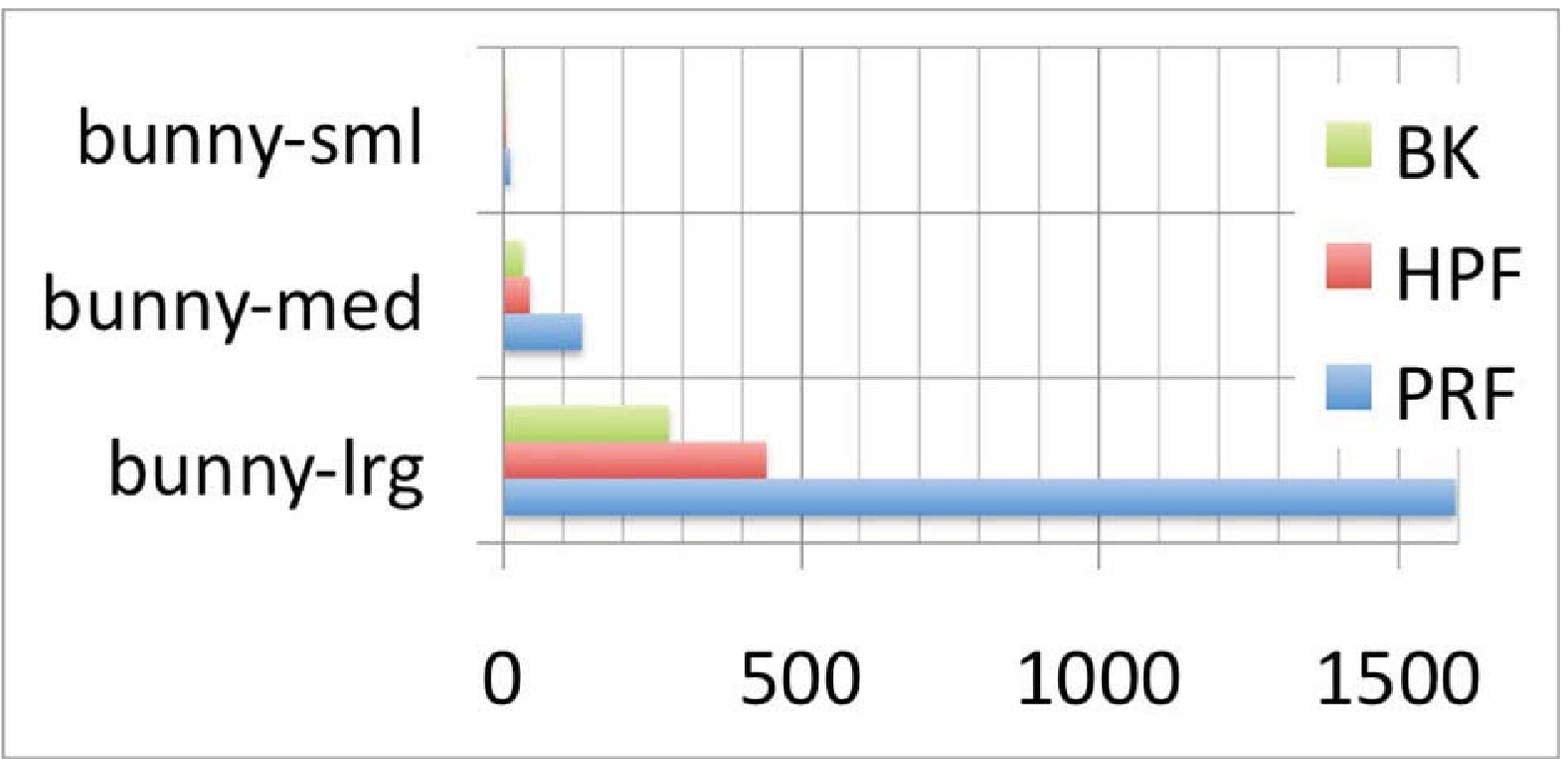}}
\label{fig:maxSurf}
}
\end{minipage}
\hspace{0.01in}
\begin{minipage}[t]{0.5\linewidth}
\centering
\subfiguretopcaptrue
\subfigure[Segmentation]{
\frame{\includegraphics[width=\linewidth]{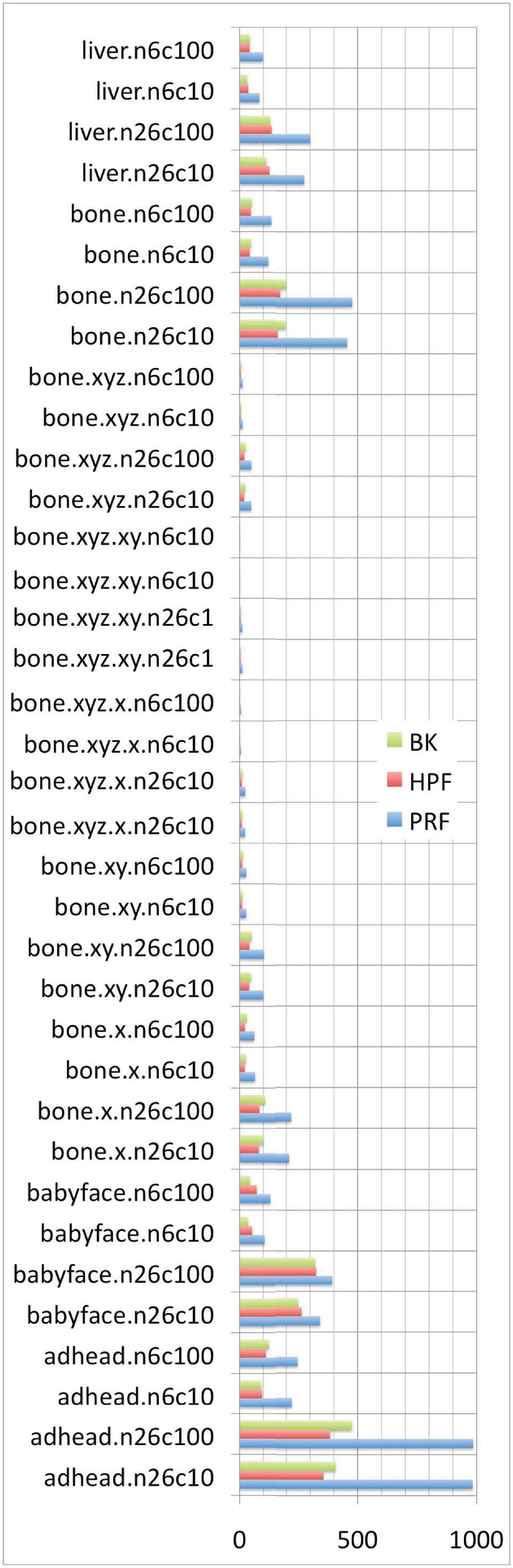}}
\label{fig:maxSeg}
}
\end{minipage}
\caption{\label{fig:maxTimes}{\bf Initialization, Minimum cut and Maximum Flow} run-times in seconds: \subref{fig:maxStereo} Stereo Problems; \subref{fig:maxMultiV} Multi-view Problems; \subref{fig:maxSurf} Surface Fitting; \subref{fig:maxSeg} Segmentation }
\end{figure}

The actual maximum-flow plays a less significant role in
solving computer vision problems. Yet, for the sake of completeness,
the maximum flow computation times of the algorithms $(t_{init} + t_{minCut} + t_{maxFlow})$ are reported in Figure \ref{fig:maxTimes} and in Tables \ref{tab:maxstereo} --
\ref{tab:maxseg}.

\subsection{Comparison to Partial Augment-Relabel}
\label{subsec:PARComp}

The PAR run-times, on our hardware setup, are deduced from the speedup factor for PAR with respect to the highest level variant of the PRF which are reported in \cite{Goldberg2008}. The paper above reports only the summation of the min-cut and max-flow run-times (without initialization), $t_{minCut} + t_{maxFlow}$ . Therefore, to enable a fair comparison we use the min-cut and max-flow run-times of the other algorithms as well. For $t_{PAR}^G$ and $t_{PRF}^G$, the run-times reported in \cite{Goldberg2008} for the PAR and PRF algorithms respectively, the estimated run-time of PAR, $\hat{t}_{PAR}$, on our hardware is: 
\begin{equation*}
\hat{t}_{PAR} = \frac{t_{PAR}^G}{t_{PRF}^G}~(t_{minCut}^{PRF} + t_{maxFlow}^{PRF})
\end{equation*}
where $t_{minCut}^{PRF}$ and $t_{maxFlow}^{PRF}$ are the corresponding run-times of the PRF algorithm measured on the hardware used in this study.

\begin{figure}[ht]

\begin{minipage}[t]{0.4\linewidth}
\centering
\subfiguretopcaptrue
\subfigure[Stereo]{
\frame{\includegraphics[width=\linewidth]{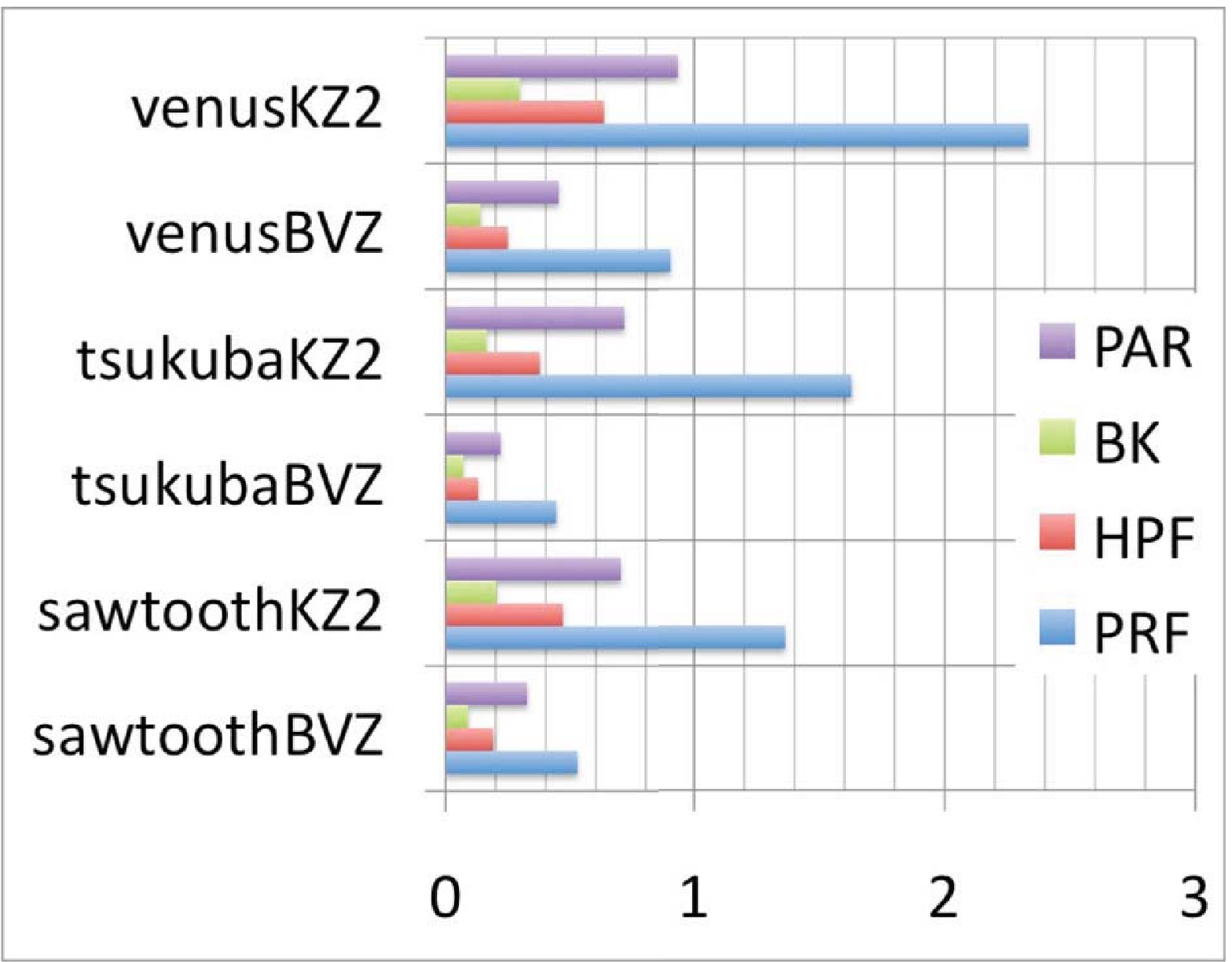}}
\label{fig:PARStereo}
}
\subfigure[Multi-View]{
\frame{\includegraphics[width=\linewidth]{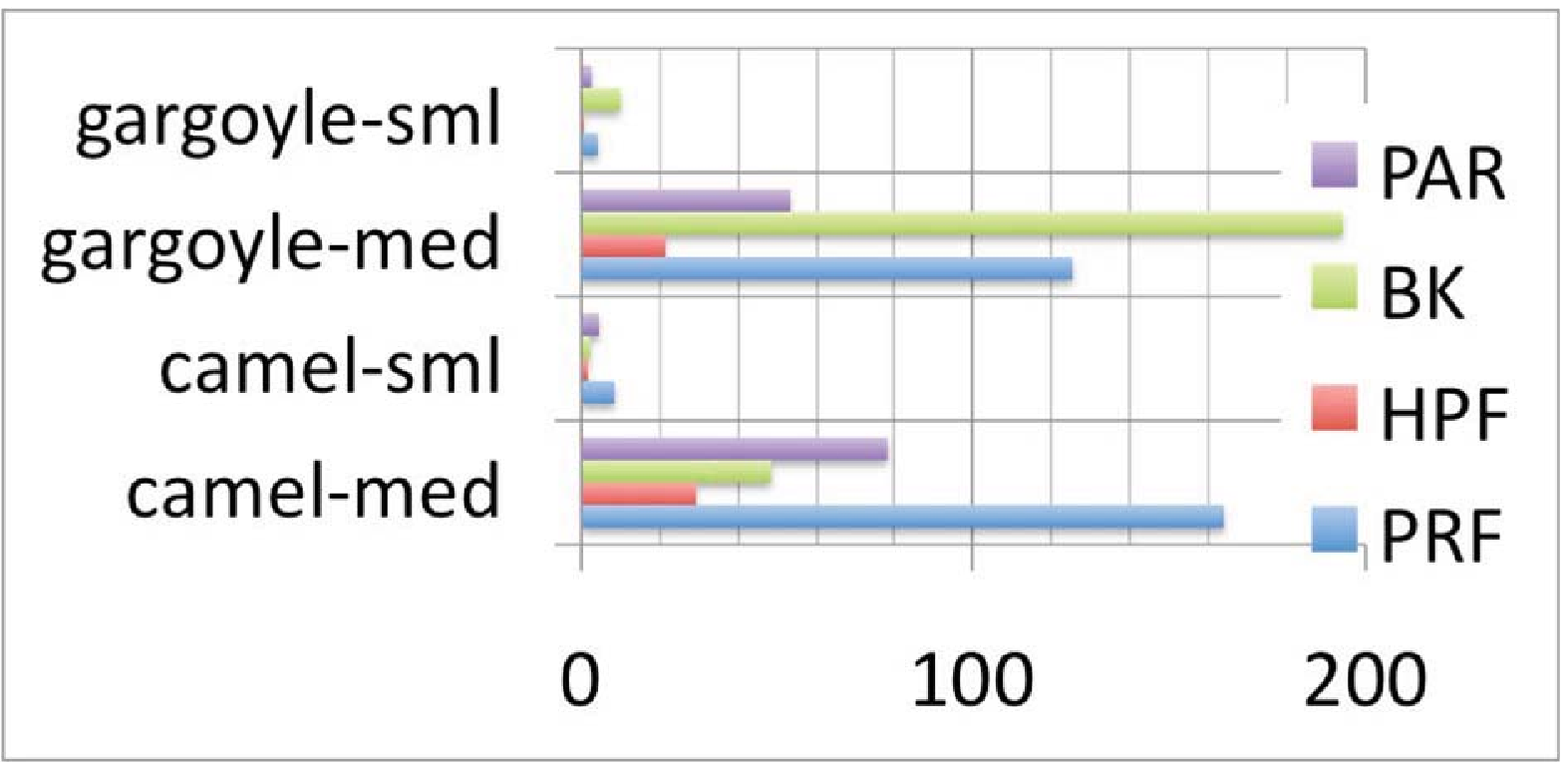}}
\label{fig:PARMultiV}
}\\
\end{minipage}
\hspace{0.01in}
\begin{minipage}[t]{0.5\linewidth}
\centering
\subfiguretopcaptrue
\subfigure[Segmentation]{
\frame{\includegraphics[width=\linewidth]{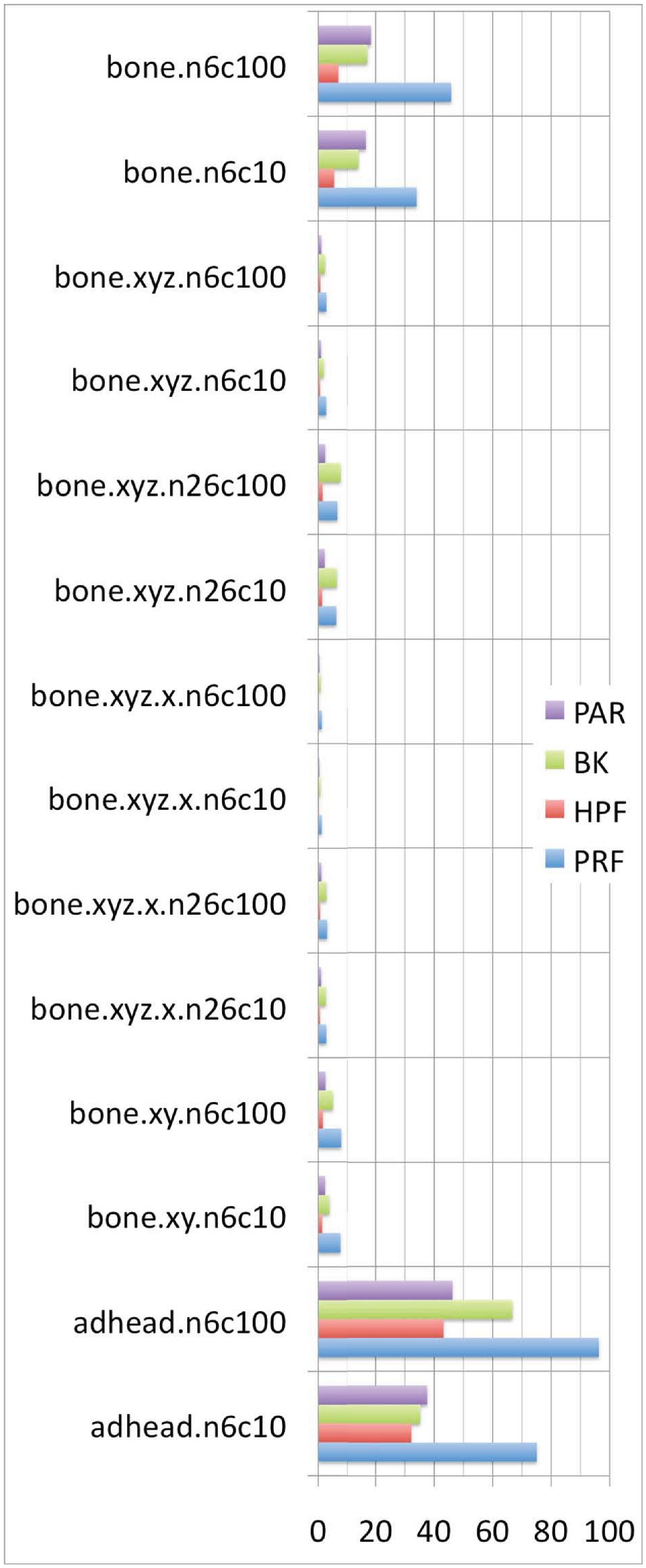}}
\label{fig:PARSeg}
}
\end{minipage}
\caption{\label{fig:PARComp} PAR, PRF, HPF, BK {\bf Minimum cut} run-times in seconds: \subref{fig:PARStereo} Stereo Problems; \subref{fig:PARMultiV} Multi-view Problems; \subref{fig:PARSeg} Segmentation}
\end{figure}

The comparison results are given in Figure \ref{fig:PARComp}
for all problem instances reported in \cite{Goldberg2008}. As
reported in \cite{Goldberg2008}, the PAR algorithm indeed
improves on PRF. HPF outperforms PAR for all problem instances. It is noted that in this comparison of the run-times that exclude initialization PAR's performance is still inferior to that of HPF.  If one were to add the initialization time then the relative performance of PAR as compare to HPF would be much worse since the initialization used has time consuming logic in it as note previously in Section \ref{sec:intro} and is shown in Figure \ref{fig:initTimes}. In terms of comparing PAR to BK, Figure \ref{fig:PARComp} shows that PAR is inferior to BK for small problem instances, but performs better for larger instances. 

\subsection{Memory Utilization}
\label{SubSec:ResultsMemory}

Measuring the actual memory utilization is of growing importance, as advances in acquisition systems and sensors allow higher image resolution, thus larger problem sizes. 

The memory utilization of each of the algorithms is a result of two factors: (i) the data each algorithm maintains in order to solve the problem. For example, the BK algorithm must maintain a flow $f$, the list of active nodes and a list of all orphans (see Section \ref{subsec:BK}); and (ii) the efficiency of the specific implementation's memory allocation in each implementation. The first factor can be analytically assessed by carefully examining the algorithm. The latter, however, must be evaluated empirically. It is important to note that both do not necessarily grow linearily with the problem size. The memory usage was read directly out of the {\em /proc/[process]/statm} file for each implementation and for each problem instance. One should note that the granularity of the information in this file is the page-file size, thus $4096$ Bytes.

\begin{figure}[!h!t!b]
  \begin{center}
\includegraphics[width=0.9\linewidth]{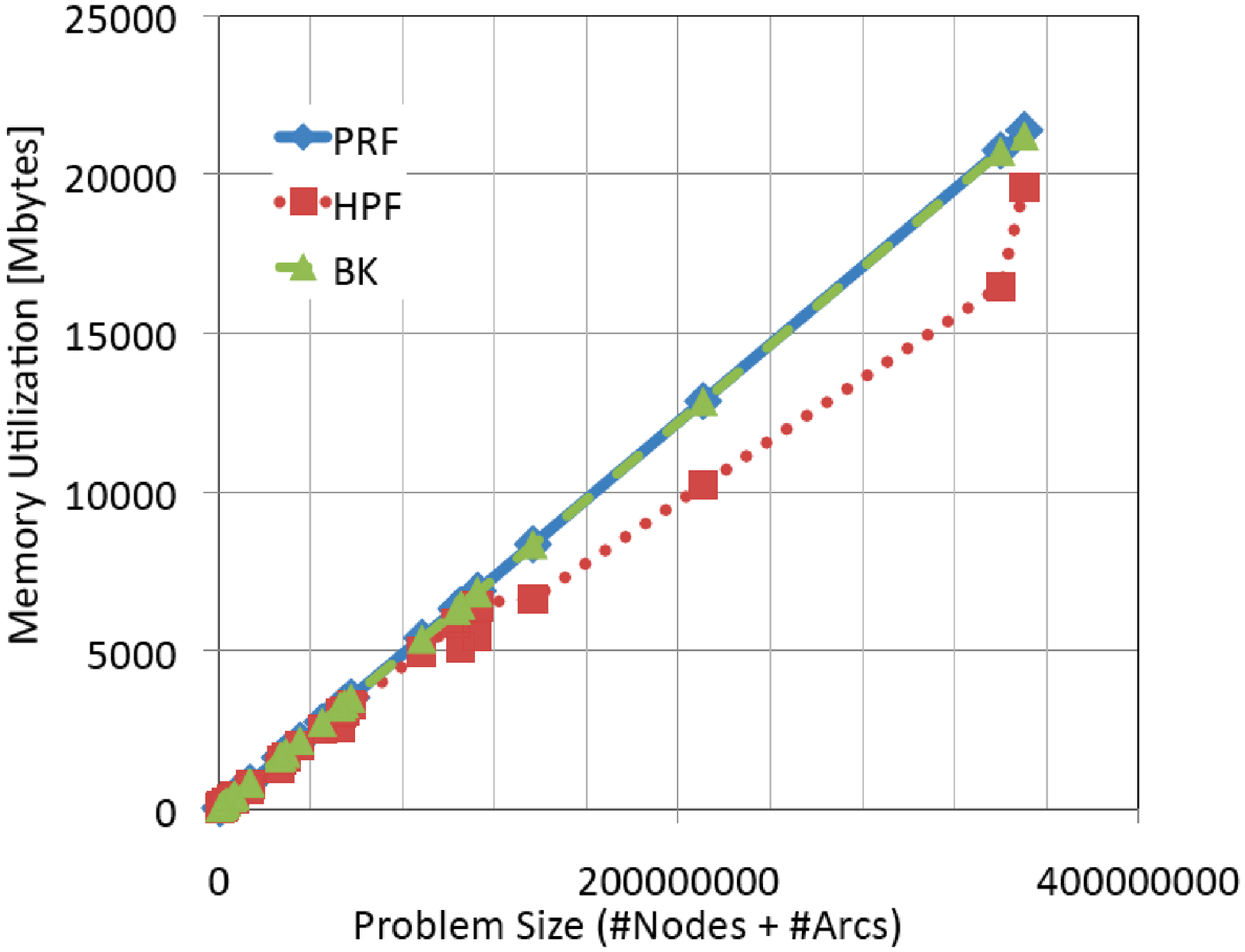}
  \end{center}
  \caption{\label{fig:memory} Memory Utilization Vs. Input size}
\end{figure}

Figure \ref{fig:memory} summarizes the results of the memory utilization for BK (solid blue line), HPF (dashed green line) and PRF (dotted red line) algorithms. These are detailed in Appendix \ref{appen:memory}, Table \ref{tab:probMemory}. The X-axis in Figure \ref{fig:memory} is the input size. A Problem's input size is the number of nodes, $n$ plus the number of arcs, $m$, in the problem's corresponding graph: $input~size = n + m$. The number of nodes, $n$, and the number of arcs, $m$, for each of the problems are given in Table \ref{tab:probSizes}. The Y-axis is the memory utilization in Mega- Bytes. 

Both BK and PRF algorithms use on average 10\% more memory than the HPF algorithm. For problem instances with large number of
arcs, the PRF and BK require 25\% more memory. This
becomes critical when the problem size is considerably large,
with respect to the machine's physical memory. In these cases
the execution of the algorithms requires a significant amount of
swapping memory pages between the physical memory and the disk,
resulting in longer execution times.

\subsection{Summary}
\label{SubSec:Summary}

\begin{figure*}[!h!t!b]
  \begin{center}
\includegraphics[width=0.7\linewidth]{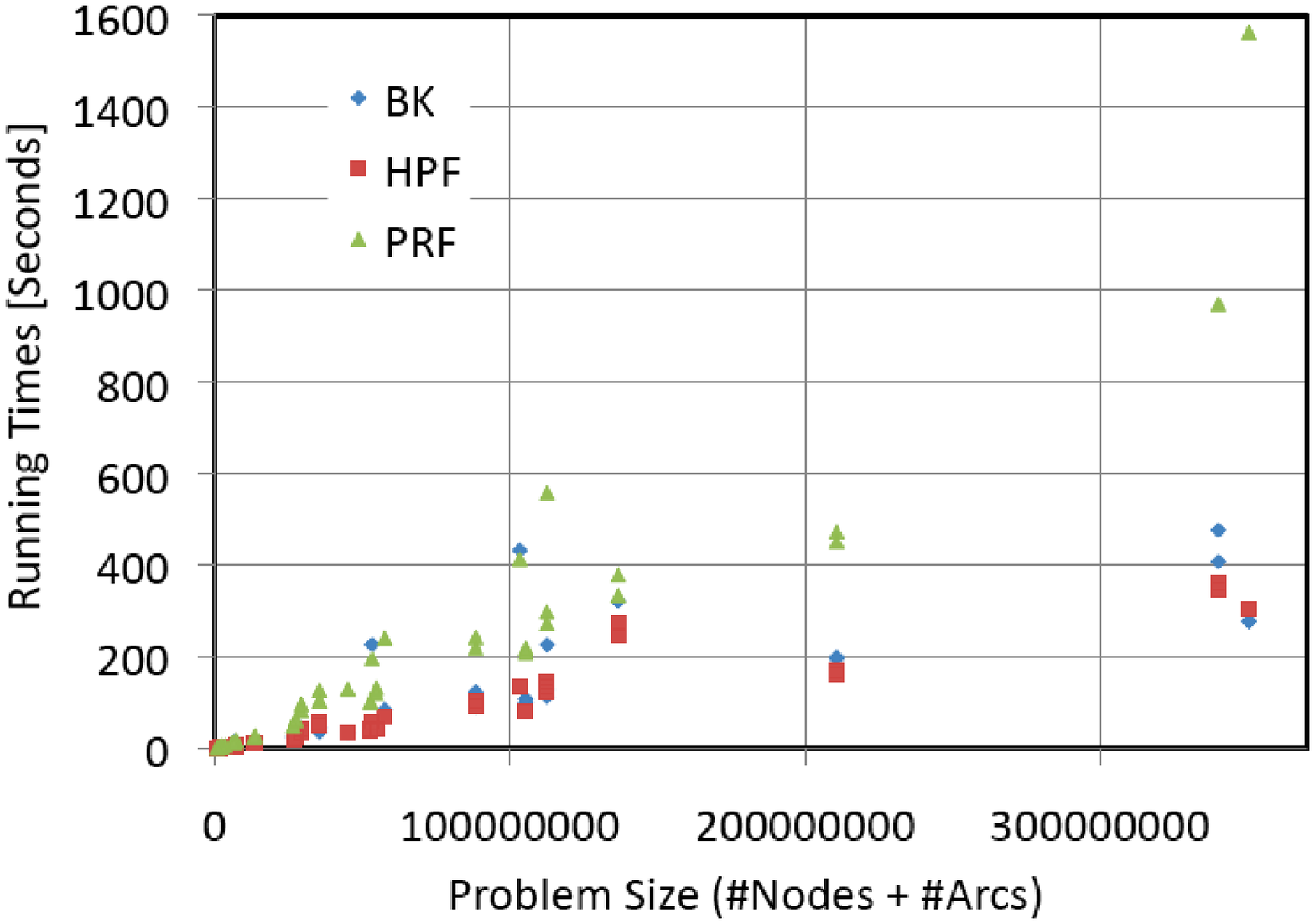}
  \end{center}
  \vspace{-0.2in}
  \caption{\label{fig:TimesSummary} Minium-cut Execution times (Initialization and minimum-cut phases) with respect to the problems' size}
\end{figure*}

Figure \ref{fig:TimesSummary} is a graphical summary of the run-times of each of the algorithms for the min-cut task ($t_{init} + t_{minCut}$) depending on the problem size. Figure \ref{fig:TimesSummary} and tables \ref{tab:probSizes} and \ref{tab:aveResults} suggest that the BK and the HPF algorithms generate comparable results. The the first, BK, is more suited for small problem instances (less then $1,000,000$ graph elements (\#Nodes + \#Arcs)) or for instances that are characterized by a short paths from $s$ to $t$. The latter, HPF, might be used for all other general larger problems. Figure \ref{fig:PARComp} shows that this also holds for PRF's revised version, the partial augment-relabel (PAR) algorithms, for all vision problem instances examined in this study. In detail, out of the $51$ instances BK is dominating $14$ times with an average running time of $48$ seconds on these instances. HPF on the other hand has an averaged running time of $54$ seconds and therefor the HPF algorithm has a slowdown factor of $1.18$ with respect to BK. On the remaining 37 instances HPF is dominating with an average running time of $72$ seconds. It takes BK $100$ seconds in average to finish on theses instances, whitch results in a slowdown factor of $1.39$ for BK with respect to HPF.  

\section{Conclusions}
\label{sec:Conclusion} This paper presents the results of a
comprehensive computational study on vision problems, in terms of execution times
and memory utilization, of four algorithms,
which optimally solve the {\em s-t} minimum cut and maximum flow
problems: (i) Goldberg's and Tarjan's {\em Push-Relabel}; (ii)
Hochbaum's {\em pseudoflow}; (iii) Boykov's and Kolmogorov's
{\em augmenting paths}; and (iv) Goldberg's {\em partial augment-relabel}. 

The results show that the BK algorithm is more suited for small problem instances (less then $1,000,000$ graph elements, thus vertices and arcs) or for instances that are characterized by short paths from $s$ to $t$, the HPF algorithm is better suited for all other general larger problems. In terms of memory utilization, the HPF algorithm has better memory utilization with up to 25\% saved in memory allocation as compared to BK and PRF.

Our results are of significance because it has been widely accepted
that both BK and PRF algorithms were the fastest algorithms in
practice for the min-cut problem. This was shown not to hold in general \cite{HPF-OR-2008}, and here for computer vision in particular. This, with the availability of HPF algorithm's source-code (see \cite{WebPS}), makes HPF the perfect tool for the growing number of computer vision applications which incorporate the min-cut problem as a sub-routine.

The current strategy of speeding up computers is to increment
the number of processors instead of increasing the computing
power of a single one. This development suggests that a
parallelization of the algorithm would be beneficial. We expect
the HPF algorithm to behave well with respect to parallel
implementations as well.

\bibliographystyle{IEEEtran}
\bibliography{Refs}

\vfill

\pagebreak
\appendices
\section{Run-times}
\label{appen:runtimes}

\begin{table}[h]
\begin{center}
\begin{tabular}{||l|rrr||}
\hline \hline
\multicolumn{4}{||l||}{{\bf Stereo}} \\ \hline
Instance & PRF & HPF & BK \\
\hline \hline
\input{InitTimesStereo.tex}
\hline \hline
\end{tabular}
\caption{\label{tab:initstereo} {\bf Initialization} stage run-times: {\em Stereo Vision} problems}
\end{center}
\end{table}

\vspace{-0.4in}
\begin{table}[h]
\begin{center}
\begin{tabular}{||l|rrr||}
\hline \hline
\multicolumn{4}{||l||}{{\bf Multi-View}} \\ \hline
Instance & PRF & HPF & BK \\
\hline \hline
\input{InitTimesMultiView.tex}
\hline \hline
\end{tabular}
\caption{\label{tab:initmultiv} {\bf Initialization} stage run-times: {\em Multi-View} problems}
\end{center}
\end{table}

\vspace{-0.4in}
\begin{table}[h]
\begin{center}
\begin{tabular}{||l|rrr||}
\hline \hline
\multicolumn{4}{||l||}{{\bf Surface Fitting}} \\ \hline
Instance & PRF & HPF & BK \\
\hline \hline
\input{InitTimesSurf.tex}
\hline \hline
\end{tabular}
\caption{\label{tab:initsurf} {\bf Initialization} stage run-times: {\em Surface Fitting} problems}
\end{center}
\end{table}

\begin{table}[h]
\begin{center}
\begin{tabular}{||l|rrr||}
\hline \hline
\multicolumn{4}{||l||}{{\bf Segmentation}} \\ \hline
Instance & PRF & HPF & BK \\
\hline \hline
\input{InitTimesSegmentation.tex}
\hline \hline
\end{tabular}
\caption{\label{tab:initseg} {\bf Initialization} stage run-times: {\em Segmentation} problems}
\end{center}
\end{table}

\vspace{-0.4in}
\begin{table}[h]
\begin{center}
\begin{tabular}{||l|rrr||}
\hline \hline
\multicolumn{4}{||l||}{{\bf Stereo}} \\ \hline
Instance & PRF & HPF & BK \\
\hline \hline
\input{maxTimesStereo.tex}
\hline \hline
\end{tabular}
\caption{\label{tab:maxstereo} Total run-times of the {\bf Initialization {\em and} Min-cut {\em and} Max-flow} stages - {\em Stereo Vision} problems}
\end{center}
\end{table}

\vspace{-0.4in}
\begin{table}[h]
\begin{center}
\begin{tabular}{||l|rrr||}
\hline \hline
\multicolumn{4}{||l||}{{\bf Multi-View}} \\ \hline
Instance & PRF & HPF & BK \\
\hline \hline
\input{maxTimesMultiV.tex}
\hline \hline
\end{tabular}
\caption{\label{tab:maxmultiv} Total run-times of the {\bf
Initialization {\em and} Min-cut {\em and} Max-flow} stages -
{\em Multi-View} problems}
\end{center}
\end{table}

\vspace{-0.4in}
\begin{table}[h]
\begin{center}
\begin{tabular}{||l|rrr||}
\hline \hline
\multicolumn{4}{||l||}{{\bf Surface Fitting}} \\ \hline
Instance & PRF & HPF & BK \\
\hline \hline
\input{maxTimesSurf.tex}

\hline \hline
\end{tabular}
\caption{\label{tab:maxsurf} Total run-times of the {\bf
Initialization {\em and} Min-cut {\em and} Max-flow} stages -
{\em Surface Fitting} problems}
\end{center}
\end{table}

\vspace{-0.4in}
\begin{table}[h]
\begin{center}
\begin{tabular}{||l|rrr||}
\hline \hline
\multicolumn{4}{||l||}{{\bf Segmentation}} \\ \hline
Instance & PRF & HPF & BK \\
\hline \hline
\input{maxTimesSeg.tex}
\hline \hline
\end{tabular}
\caption{\label{tab:maxseg} Total run-times of the {\bf
Initialization {\em and} Min-cut {\em and} Max-flow} stages -
{\em Segmentation} problems}
\end{center}
\end{table}

\FloatBarrier

\section{Memory Utilization}
\label{appen:memory}

\begin{table}[!h!t!b]
\begin{center}
\footnotesize
\begin{tabular}{||l|ccc||}
\hline \hline
{Instance} & {PRF} & {HPF} &{BK}\\
\hline \hline
\multicolumn{4}{||l||}{{\bf Stereo}} \\ \hline
\input{memory_3_algs.tex}
\hline \hline
\end{tabular}
\caption{\label{tab:probMemory} Memory Utilization in [MBytes]}
\end{center}
\end{table}

\ifCLASSOPTIONcaptionsoff
  \newpage
\fi

\end{document}

%% file: algo.tex
\newcounter{gap}
\newcommand{\nn}{

\noindent \hspace{\thegap em} }

\newcommand{\Procedure}{\noindent {\bf procedure }\addtocounter{gap}{2}}

\newcommand{\If}{{\bf if }}

\newcommand{\Begin}{{\bf begin}  \addtocounter{gap}{2}}

\newcommand{\Else}{{\bf else }}

\newcommand{\Do}{\addtocounter{gap}{2}{\bf do}}

\newcommand{\While}{{\bf while }}

\newcommand{\End}{{\bf end}}

\newcommand{\Comment}[1]{
{\it
/*\\#1\\ */} \medskip

}

\newcommand{\EndLoop}{\addtocounter{gap}{-2}}

%% file: table1ratio.tex
sawtoothBVZ	&173,602	&838,635	&1.55	&0.76	&\textbf{0.62} & 2.50 & 1.23 & \textbf{1} \\
sawtoothKZ2	&310,459	&2,059,153	&4.05	&1.87	&\textbf{1.52} & 2.66 & 1.23 & \textbf{1} \\
tsukubaBVZ	&117,967	&547,699	&1.09	&0.50	&\textbf{0.41} & 2.66 & 1.22 & \textbf{1} \\
tsukubaKZ2	&213,144	&1,430,508	&3.42	&1.33	&\textbf{1.06} & 3.23 & 1.25 & \textbf{1} \\
venusBVZ	&174,139	&833,168	&1.93	&0.82	&\textbf{0.66} & 2.92 & 1.24 & \textbf{1} \\
venusKZ2	&315,972	&2,122,772	&5.12	&2.09	&\textbf{1.66} & 3.08 & 1.26 & \textbf{1}\\
\hline
\multicolumn{9}{||l||}{{\bf Multi-View}} \\ \hline
camel-lrg	&18,900,002	&93,749,846	&558.65	&\textbf{143.96}	&225.53 & 3.88 & \textbf{1} & 1.57 \\
camel-med	&9,676,802	&47,933,324	&240.30	&\textbf{67.61}	&83.43 & 3.55 & \textbf{1} & 1.23\\
camel-sml	&1,209,602	&5,963,582	&17.16	&\textbf{6.31}	&6.46 & 2.72 & \textbf{1} & 1.02\\
gargoyle-lrg	&17,203,202	&86,175,090	&413.26	&\textbf{134.77}	&432.31 & 3.07 & \textbf{1} & 3.21\\
gargoyle-med	&8,847,362	&44,398,548	&196.26	&\textbf{56.24}	&226.43 & 3.49 & \textbf{1} & 4.03 \\
gargoyle-sml	&1,105,922	&5,604,568	&12.26	&\textbf{5.00}	&13.86 & 2.45 & \textbf{1} & 2.77\\
\hline
\multicolumn{9}{||l||}{{\bf Surface Fitting}} \\ \hline
bunny-lrg	&49,544,354	&300,838,741	&1564.10	&305.05	&\textbf{277.02} & 5.56 & 1.10 & \textbf{1}\\
bunny-med	&6,311,088	&38,739,041	&129.05	&34.47	&\textbf{32.82} & 3.93 & 1.05 & \textbf{1}\\
bunny-sml	&805,802	&5,040,834	&10.89	&4.12	&\textbf{4.03} & 2.70 & 1.02 & \textbf{1}\\
\hline
\multicolumn{9}{||l||}{{\bf Segmentation}} \\ \hline
adhead.n26c10	&12,582,914	&327,484,556	&970.74	&\textbf{344.31}	&407.42 & 2.82	 & \textbf{1} &	1.18\\
adhead.n26c100	&12,582,914	&327,484,556	&971.20	&\textbf{362.49}	&476.04 & 2.68	 & \textbf{1} & 1.31 \\
adhead.n6c10	&12,582,914	&75,826,316	&219.14	&\textbf{90.17}	&90.38 & 2.43 & \textbf{1} &	1.01\\
adhead.n6c100	&12,582,914	&75,826,316	&242.43	&\textbf{103.03}	&123.22 & 2.35	 & \textbf{1} & 1.20\\
babyface.n26c10	&5,062,502	&131,636,370	&333.91	&\textbf{245.10} &250.15 & 1.36	 & \textbf{1}	& 1.02
\\
babyface.n26c100	&5,062,502	&131,636,370	&378.94	&\textbf{272.26}	&321.20 & 1.39	 & \textbf{1}	& 1.18\\
babyface.n6c10	&5,062,502	&30,386,370	&103.27	&48.30	&\textbf{35.30} & 2.93	& 1.37 &	\textbf{1}\\
babyface.n6c100	&5,062,502	&30,386,370	&126.28	&58.77	&\textbf{43.89} & 2.88	& 1.34 &	\textbf{1}\\
bone.n26c10	&7,798,786	&202,895,861	&451.35	&\textbf{160.73}	&196.26 & 2.81	& \textbf{1}	& 1.22\\
bone.n26c100	&7,798,786	&202,895,861	&472.34	&\textbf{168.09}	&198.73 & 2.81	& \textbf{1}	& 1.18\\
bone.n6c10	&7,798,786	&46,920,181	&119.76	&\textbf{41.02}	&47.79 & 2.92	& \textbf{1}	& 1.17\\
bone.n6c100	&7,798,786	&46,920,181	&132.06	&\textbf{43.56}	&51.88 & 3.03	& \textbf{1}	& 1.19\\
bone\_subx.n26c10	&3,899,394	&101,476,818	&209.03	&\textbf{79.70}	&100.23 & 2.62	 & \textbf{1} &	1.26\\
bone\_subx.n26c100	&3,899,394	&101,476,818	&218.57	&\textbf{81.79}	&107.43 & 2.67	& \textbf{1} &	1.31\\
bone\_subx.n6c10	&3,899,394	&23,488,978	&64.17	&\textbf{20.89}	&26.57 & 3.07	& \textbf{1} &	1.27\\
bone\_subx.n6c100	&3,899,394	&23,488,978	&61.18	&\textbf{22.06}	&30.29 & 2.77	& \textbf{1} &	1.37\\
bone\_subxy.n26c10	&1,949,698	&50,753,434	&99.59	&\textbf{39.51}	&48.25 & 2.52	& \textbf{1} &	1.22\\
bone\_subxy.n26c100	&1,949,698	&50,753,434	&101.13	&\textbf{40.00}	&51.04 & 2.53	& \textbf{1} &	1.28\\
bone\_subxy.n6c10	&1,949,698	&11,759,514	&27.22	&\textbf{10.04}	&12.09 & 2.71	& \textbf{1} &	1.20\\
bone\_subxy.n6c100	&1,949,698	&11,759,514	&27.66	&\textbf{10.56}	&13.62 & 2.62	& \textbf{1} &	1.29\\
bone\_subxyz.n26c10	&983,042	&25,590,293	&48.07	&\textbf{18.89}	&23.69 & 2.54	& \textbf{1} &	1.25\\
bone\_subxyz.n26c100	&983,042	&25,590,293	&48.83	&\textbf{19.39}	&25.41 & 2.52	& \textbf{1} &	1.31\\
bone\_subxyz.n6c10	&983,042	&5,929,493	&11.95	&\textbf{4.85}	&5.95 & 2.46 &	\textbf{1} &	1.23\\
bone\_subxyz.n6c100	&983,042	&5,929,493	&12.05	&\textbf{5.14}	&6.54 & 2.34	& \textbf{1} &	1.27\\
bone\_subxyz\_subx.n26c10	&491,522	&12,802,789	&22.96	&\textbf{9.36}	&11.27 & 2.45	& \textbf{1} &	1.20\\
bone\_subxyz\_subx.n26c100	&491,522	&12,802,789	&23.55	&\textbf{9.52}	&11.55 & 2.47	& \textbf{1} &	1.21\\
bone\_subxyz\_subx.n6c10	&491,522	&2,972,389	&5.47	&\textbf{2.37}	&2.75 & 2.31 &	\textbf{1} &	1.16\\
bone\_subxyz\_subx.n6c100	&491,522	&2,972,389	&5.56	&\textbf{2.47}	&2.89 & 2.25 &	\textbf{1}	& 1.17 \\
bone\_subxyz\_subxy.n26c10	&245,762	&6,405,104	&10.99	&\textbf{4.63}	&5.60 & 2.37 &	\textbf{1}	& 1.21 \\
bone\_subxyz\_subxy.n26c100	&245,762	&6,405,104	&11.06	&\textbf{4.74}	&5.79 & 2.33 &	\textbf{1}	& 1.22 \\
bone\_subxyz\_subxy.n6c10	&245,762	&1,489,904	&2.55	&\textbf{1.14}	&1.34 & 2.24 &	\textbf{1}	& 1.18 \\
bone\_subxyz\_subxy.n6c100	&245,762	&1,489,904	&2.59	&\textbf{1.21}	&1.39 & 2.14 &	\textbf{1}	& 1.15 \\
liver.n26c10	&4,161,602	&108,370,821	&272.63	&123.61	&\textbf{112.40} & 2.43 &	1.10	& \textbf{1} \\
liver.n26c100	&4,161,602	&108,370,821	&297.88	&132.48	&\textbf{128.60} & 2.32 &	1.03	& \textbf{1} \\
liver.n6c10	&4,161,602	&25,138,821	&82.11	&35.24	&\textbf{32.02} & 2.56 &  	1.10	& \textbf{1} \\
liver.n6c100	&4,161,602	&25,138,821	&96.38	&\textbf{40.36}	&43.60 & 2.39	& \textbf{1} &	1.08 \\

%% file: InitTimesStereo.tex
sawtoothBVZ	&1.02	&0.57	&0.53\\
sawtoothKZ2	&2.69	&1.40	&1.32\\
tsukubaBVZ	&0.65	&0.37	&0.34\\
tsukubaKZ2	&1.79	&0.96	&0.90\\
venusBVZ	&1.02	&0.57	&0.53\\
venusKZ2	&2.79	&1.45	&1.36\\
\hline
{\bf Average}	&{\bf 1.66}	&{\bf 0.89}	&{\bf 0.83}\\

%% file: InitTimesMultiView.tex
camel-lrg	&155.25	&76.97	&70.22\\
camel-med	&76.40	&38.44	&35.15\\
camel-sml	&8.66	&4.65	&4.23\\
gargoyle-lrg	&141.81	&68.79	&63.50\\
gargoyle-med	&70.79	&34.97	&32.33\\
gargoyle-sml	&8.11	&4.33	&3.94\\
\hline
{\bf Average}	&{\bf 76.84}	&{\bf 38.02}	&{\bf 34.89}\\

%% file: InitTimesSurf.tex
bunny-lrg	&687.01	&230.81	&219.08\\
bunny-med	&70.87	&29.25	&27.95\\
bunny-sml	&7.99	&3.70	&3.47\\
\hline
{\bf Average}	&{\bf 255.29}	&{\bf 87.92}	&{\bf 83.50}\\

%% file: InitTimesSegmentation.tex
adhead.n26c10	&697.57	&238.71	&233.34\\
adhead.n26c100	&702.45	&242.09	&239.02\\
adhead.n6c10	&144.05	&58.00	&55.12\\
adhead.n6c100	&146.12	&59.82	&56.47\\
babyface.n26c10	&180.35	&94.24	&92.50\\
babyface.n26c100	&182.76	&95.88	&94.62\\
babyface.n6c10	&44.31	&22.77	&21.71\\
babyface.n6c100	&44.85	&23.37	&22.35\\
bone.n26c10	&381.60	&146.00	&144.64\\
bone.n26c100	&382.59	&149.93	&145.70\\
bone.n6c10	&85.74	&35.58	&33.66\\
bone.n6c100	&86.26	&36.68	&34.72\\
bone\_subx.n26c10	&182.47	&72.52	&71.59\\
bone\_subx.n26c100	&183.88	&73.70	&72.54\\
bone\_subx.n6c10	&41.02	&17.62	&16.78\\
bone\_subx.n6c100	&41.46	&18.18	&17.28\\
bone\_subxy.n26c10	&87.33	&36.05	&35.32\\
bone\_subxy.n26c100	&88.08	&36.37	&35.89\\
bone\_subxy.n6c10	&19.62	&8.73	&8.26\\
bone\_subxy.n6c100	&19.77	&8.99	&8.60\\
bone\_subxyz.n26c10	&41.91	&17.60	&17.31\\
bone\_subxyz.n26c100	&42.26	&17.93	&17.74\\
bone\_subxyz.n6c10	&9.26	&4.28	&4.10\\
bone\_subxyz.n6c100	&9.29	&4.46	&4.21\\
bone\_subxyz\_subx.n26c10	&20.24	&8.80	&8.66\\
bone\_subxyz\_subx.n26c100	&20.47	&8.93	&8.76\\
bone\_subxyz\_subx.n6c10	&4.32	&2.16	&2.05\\
bone\_subxyz\_subx.n6c100	&4.39	&2.23	&2.10\\
bone\_subxyz\_subxy.n26c10	&9.63	&4.38	&4.30\\
bone\_subxyz\_subxy.n26c100	&9.68	&4.47	&4.34\\
bone\_subxyz\_subxy.n6c10	&2.05	&1.06	&1.02\\
bone\_subxyz\_subxy.n6c100	&2.08	&1.11	&1.05\\
liver.n26c10	&200.28	&76.43	&75.65\\
liver.n26c100	&200.64	&77.25	&76.02\\
liver.n6c10	&44.58	&18.68	&17.90\\
liver.n6c100	&44.76	&18.91	&17.99\\
\hline
{\bf Average}	&{\bf 122.45}	&{\bf 48.44}	&{\bf 47.31}\\

%% file: maxTimesStereo.tex
sawtoothBVZ	&1.64	&0.88	&0.62\\
sawtoothKZ2	&4.26	&2.02	&1.52\\
tsukubaBVZ	&1.15	&0.57	&0.41\\
tsukubaKZ2	&3.55	&1.42	&1.06\\
venusBVZ	&2.02	&0.94	&0.66\\
venusKZ2	&5.32	&2.22	&1.66\\
\hline
{\bf Average}	&{\bf 2.99}	&{\bf 1.34}	&{\bf 0.99}\\

%% file: maxTimesMultiV.tex
camel-lrg	&563.52	&180.57	&225.53\\
camel-med	&242.74	&80.56	&83.43\\
camel-sml	&17.44	&6.92	&6.46\\
gargoyle-lrg	&417.75	&154.40	&432.31\\
gargoyle-med	&198.56	&66.66	&226.43\\
gargoyle-sml	&12.54	&5.61	&13.86\\
\hline
{\bf Average}	&{\bf 242.09}	&{\bf 82.45}	&{\bf 164.67}\\

%% file: maxTimesSurf.tex
bunny-lrg	&1595.21	&440.59	&277.02\\
bunny-med	&131.78	&43.50	&32.82\\
bunny-sml	&11.18	&4.74	&4.03\\
\hline
{\bf Average}	&{\bf 579.39}	&{\bf 162.94}	&{\bf 104.62}\\

%% file: maxTimesSeg.tex
adhead.n26c10	&982.98	&356.07	&407.42\\
adhead.n26c100	&985.39	&383.55	&476.04\\
adhead.n6c10	&223.30	&94.54	&90.38\\
adhead.n6c100	&248.06	&110.58	&123.22\\
babyface.n26c10	&341.32	&264.41	&250.15\\
babyface.n26c100	&392.62	&325.34	&321.20\\
babyface.n6c10	&105.48	&52.74	&35.30\\
babyface.n6c100	&129.39	&72.35	&43.89\\
bone.n26c10	&456.03	&163.58	&196.26\\
bone.n26c100	&477.26	&174.07	&198.73\\
bone.n6c10	&121.31	&42.74	&47.79\\
bone.n6c100	&133.96	&47.02	&51.88\\
bone\_subx.n26c10	&211.19	&80.94	&100.23\\
bone\_subx.n26c100	&220.91	&83.42	&107.43\\
bone\_subx.n6c10	&64.91	&21.41	&26.57\\
bone\_subx.n6c100	&62.00	&22.52	&30.29\\
bone\_subxy.n26c10	&100.65	&40.13	&48.25\\
bone\_subxy.n26c100	&102.25	&40.87	&51.04\\
bone\_subxy.n6c10	&27.58	&10.30	&12.09\\
bone\_subxy.n6c100	&28.05	&10.83	&13.62\\
bone\_subxyz.n26c10	&48.60	&19.15	&23.69\\
bone\_subxyz.n26c100	&49.39	&19.71	&25.41\\
bone\_subxyz.n6c10	&12.13	&4.98	&5.95\\
bone\_subxyz.n6c100	&12.24	&5.28	&6.54\\
bone\_subxyz\_subx.n26c10	&23.23	&9.48	&11.27\\
bone\_subxyz\_subx.n26c100	&23.82	&9.68	&11.55\\
bone\_subxyz\_subx.n6c10	&5.56	&2.43	&2.75\\
bone\_subxyz\_subx.n6c100	&5.65	&2.54	&2.89\\
bone\_subxyz\_subxy.n26c10	&11.12	&4.67	&5.60\\
bone\_subxyz\_subxy.n26c100	&11.20	&4.80	&5.79\\
bone\_subxyz\_subxy.n6c10	&2.60	&1.16	&1.34\\
bone\_subxyz\_subxy.n6c100	&2.64	&1.24	&1.39\\
liver.n26c10	&275.68	&126.03	&112.40\\
liver.n26c100	&301.11	&135.20	&128.60\\
liver.n6c10	&83.42	&36.99	&32.02\\
liver.n6c100	&97.88	&42.30	&43.60\\
\hline
{\bf Average}	&{\bf 177.25}	&{\bf 78.42}	&{\bf 84.79}\\

%% file: memory_3_algs.tex
sawtoothBVZ	&62.28	&58.50	&69.27\\
sawtoothKZ2	&141.08	&125.81	&147.55\\
tsukubaBVZ	&41.62	&39.55	&48.79\\
tsukubaKZ2	&97.72	&87.12	&104.57\\
venusBVZ	&62.27	&58.65	&69.24\\
venusKZ2	&145.76	&129.76	&152.22\\
\hline
\multicolumn{4}{||l||}{{\bf Segmentation}} \\ \hline
adhead.n26c10	&20,759.80	&16,480.20	&20,719.40\\
adhead.n26c100	&20,759.80	&16,480.20	&20,719.40\\
adhead.n6c10	&5,399.80	&4,960.20	&5,359.40\\
adhead.n6c100	&5,399.80	&4,960.20	&5,359.40\\
babyface.n26c10	&8,347.10	&6,628.00	&8,335.40\\
babyface.n26c100	&8,347.10	&6,628.00	&8,335.40\\
babyface.n6c10	&2,167.30	&1,993.20	&2,155.60\\
babyface.n6c100	&2,167.30	&1,993.20	&2,155.60\\
bone.n26c10	&12,863.50	&10,212.70	&12,841.30\\
bone.n26c100	&12,863.50	&10,212.70	&12,841.30\\
bone.n6c10	&3,343.50	&3,072.70	&3,321.30\\
bone.n6c100	&3,343.50	&3,072.70	&3,321.30\\
bone\_subx.n26c10	&6,435.30	&5,109.30	&6,428.10\\
bone\_subx.n26c100	&6,435.30	&5,109.30	&6,428.10\\
bone\_subx.n6c10	&1,675.30	&1,539.30	&1,668.10\\
bone\_subx.n6c100	&1,675.30	&1,539.30	&1,668.10\\
bone\_subxy.n26c10	&3,220.40	&2,557.04	&3,220.60\\
bone\_subxy.n26c100	&3,220.40	&2,557.06	&3,220.60\\
bone\_subxy.n6c10	&840.40	&772.00	&840.60\\
bone\_subxy.n6c100	&840.40	&772.00	&840.60\\
bone\_subxyz.n26c10	&1,625.60	&1,291.00	&1,629.40\\
bone\_subxyz.n26c100	&1,625.60	&1,291.00	&1,629.40\\
bone\_subxyz.n6c10	&425.60	&391.10	&429.40\\
bone\_subxyz.n6c100	&425.60	&391.10	&429.40\\
bone\_subxyz\_subx.n26c10	&815.10	&647.60	&820.80\\
bone\_subxyz\_subx.n26c100	&815.10	&647.60	&820.80\\
bone\_subxyz\_subx.n6c10	&215.10	&197.70	&220.80\\
bone\_subxyz\_subx.n6c100	&215.10	&197.70	&220.80\\
bone\_subxyz\_subxy.n26c10	&409.60	&325.80	&416.30\\
bone\_subxyz\_subxy.n26c100	&409.60	&325.80	&416.30\\
bone\_subxyz\_subxy.n6c10	&109.60	&100.80	&116.30\\
bone\_subxyz\_subxy.n6c100	&109.60	&100.80	&116.30\\
liver.n26c10	&6,872.10	&5,455.20	&6,863.80\\
liver.n26c100	&6,872.10	&5,455.20	&6,863.80\\
liver.n6c10	&1,792.00	&1,645.20	&1,783.80\\
liver.n6c100	&1,792.00	&1,645.20	&1,783.80\\
\hline
\multicolumn{4}{||l||}{{\bf Multi-View}} \\ \hline
camel-lrg	&6,879.30	&6,392.60	&6,814.80\\
camel-med	&3,519.90	&3,272.50	&3,490.60\\
camel-sml	&441.50	&411.30	&444.50\\
gargoyle-lrg	&6,313.40	&5,851.80	&6,255.40\\
gargoyle-med	&3,253.60	&3,015.00	&3,227.40\\
gargoyle-sml	&413.30	&382.52	&416.60\\
\hline
\multicolumn{4}{||l||}{{\bf Surface Fitting}} \\ \hline
bunny-lrg	&21,389.40	&19,600.30	&21,208.00\\
bunny-med	&2,753.30	&2,515.00	&2,736.80\\
bunny-sml	&360.50	&327.70	&365.10\\